\documentclass[conference]{IEEEtran}

\makeatletter
\let\NAT@parse\undefined
\makeatother
\usepackage[numbers]{natbib}

\usepackage{moreverb,url}

\usepackage[colorlinks,bookmarksopen,bookmarksnumbered,citecolor=red,urlcolor=red]{hyperref}

\newcommand\BibTeX{{\rmfamily B\kern-.05em \textsc{i\kern-.025em b}\kern-.08em
T\kern-.1667em\lower.7ex\hbox{E}\kern-.125emX}}

\usepackage{tikz}
\usepackage{pgfplots}
\usepackage{pgfplotstable}
\usepgfplotslibrary{groupplots}
\usepackage{subcaption}
\usepackage{multirow}
\usepackage{amsmath}
\usepackage{endnotes} 
\usepackage{booktabs}

\definecolor{mpi-green}{RGB}{0,117,103}
\definecolor{mpi-grey}{RGB}{211,211,205}
\definecolor{mpi-lblue}{RGB}{58,185,198}
\definecolor{mpi-blue}{RGB}{45,147,207}
\definecolor{mpi-red}{RGB}{204,75,99}
\definecolor{mpi-lgreen}{RGB}{75,168,79}
\definecolor{mpi-purple}{RGB}{114,122,178}
\definecolor{mpi-orange}{RGB}{247,179,94}
\definecolor{plot-orange}{RGB}{229,133,11}
\definecolor{plot-grey}{RGB}{182,183,175}

\begin{document}

%\runninghead{Kloss et al.}

\title{Combining Learned and Analytical Models for Predicting Action Effects from Sensory Data}

%\author{Alina Kloss\affilnum{1}, Stefan Schaal\affilnum{2} and Jeannette Bohg\affilnum{3}}

%\affiliation{\affilnum{1}Autonomous Motion Department, MPI for Intelligent Systems, Germany \\
%\affilnum{2}Computational Learning and Motor Control Lab, University of Southern California, USA \\
%\affilnum{3}Department of Computer Science, Stanford University, USA}

%\corrauth{Alina Kloss, Autonomous Motion Department, 
%MPI for Intelligent Systems, 
%Max-Planck-Ring 4,
%72076 T\"ubingen,
%Germany}

%\email{alina.kloss@tuebingen.mpg.de}

\author{\authorblockN{Alina Kloss\authorrefmark{1},
Stefan Schaal\authorrefmark{2}\authorrefmark{1},
and
Jeannette Bohg\authorrefmark{3}}
\authorblockA{\authorrefmark{1}Autonomous Motion Department, Max Planck Institute for Intelligent Systems, Germany }
%			   \\
%				Email: first.lastname@tue.mpg.de}
\authorblockA{\authorrefmark{2}Computational Learning and Motor Control Lab, University of Southern California, USA}
\authorblockA{\authorrefmark{3}Department of Computer Science, Stanford University, USA}
}

%\keywords{}

\maketitle

\begin{abstract}
One of the most basic skills a robot should possess is predicting 
the effect of physical interactions with objects in the environment. 
This enables optimal action selection to reach a certain goal state. 
Traditionally, dynamics are approximated by physics-based analytical models.
These models rely on specific state representations that may be hard to obtain from 
raw sensory data, especially if no knowledge of the object shape is assumed. 
More recently, we have seen learning approaches that can predict the effect 
of complex physical interactions directly from sensory input. It is 
however an open question how far these models generalize beyond their 
training data. In this work, we investigate the advantages and 
limitations of neural network based learning approaches for predicting the effects
of actions based on sensory input and show how 
analytical and learned models can be combined to leverage the best of 
both worlds. As physical interaction task, we use planar pushing, for 
which there exists a well-known analytical model {\em and\/} a large 
real-world dataset.
We propose to use a convolutional neural network to convert raw depth images or
organized point clouds into 
a suitable representation for the analytical model and compare this approach 
to using neural networks for both, perception and prediction. 
A systematic evaluation of the
proposed approach on a very large real-world dataset shows two
main advantages of the hybrid architecture. Compared to a pure neural 
network, it significantly (i) reduces required training data and (ii) 
improves generalization to novel physical interaction.  
\end{abstract}

\section{Introduction}
\label{sec:introduction}

We approach the problem of predicting the consequences of physical
interaction with objects in the environment based on raw sensory data. 
Traditionally, interaction dynamics are described by a physics-based 
analytical model~\citep{data,model,6225125} which relies on a certain 
representation of the environment state. This 
approach has the advantage that the underlying function and the input
parameters to the model have physical meaning and can therefore be 
transferred to problems with variations of these parameters. They also
make the underlying assumptions in the model transparent. However,
defining such models for complex scenarios and extracting the
required state representation from raw sensory data may be very hard,
especially if no assumptions about the shape of objects are made. 

More recently, we have seen approaches that successfully replace the 
physics-based models with learned ones
\citep*{zhou, belter2014, mericcli, kopicki2017, bauza}. While 
often more accurate than analytical models, these methods still 
assume a predefined state representation as input and do not address
the problem of how it may be extracted from raw sensory data.

Some neural network based methods instead simultaneously learn 
a representation of the input and the associated dynamics 
from large amounts of training data, e.g.\ \citep{se3, 
poke,visual_interaction,NIPS2016_6161}. They have shown impressive 
results in predicting the effect of physical interactions. In 
\citep{poke}, the authors argue that a neural network may benefit from 
choosing its own representation of the input data instead of being 
forced to use a predefined state representation. They reason that a
problem can often be parametrized in different ways and that some of
these parametrizations might be easier to obtain from the given sensory 
input than others.
The disadvantage of a learned representation is however that it usually cannot
be be mapped to physical quantities. This makes it hard to intuitively
understand the learned functions and representations. 
%It also makes it
%hard to find some guarantees or to establish bounds on the
%input parameters to the dynamics model. 
In addition, it remains unclear how these models could 
be transferred to similar problems. Neural networks often 
have the capacity to memorize their training data~\citep{generalization} 
and learn a mapping from inputs to outputs instead of the ``true''
underlying function. This can make perfect sense if the 
training data covers the whole problem domain. However, when data is sparse
(e.g.\ because a robot learns by experimenting), the question of how to 
generalize beyond the training data becomes very important.

Our hypothesis is that using prior knowledge from existing physics-based
models can provide a way to reduce the amount of required training data 
and at the same time ensure good generalization beyond the training domain.  
In this paper, we thus investigate using neural networks for
extracting a suitable representation from raw sensory data that can then
be consumed by an analytical model for prediction. We compare this
hybrid approach to using a neural network for both perception and
prediction and to the analytical model applied on ground truth input values.

As example physical interaction task, we choose planar pushing. For
this task, a well-known physical model~\citep{model} is available as
well as a large, real-world dataset~\citep{data} which we augmented with
simulated images. Given a depth image of a tabletop scene with one
object and the position and movement of the pusher, our
models need to predict the object position in the given image and
its movement due to the push. Although the state-space of the object is rather 
low-dimensional (2D position plus orientation), pushing is already a quite 
complex manipulation problem: The system is under-actuated and the 
relationship between the push and the object movement is highly non-linear. 
The pusher can slide along the object and dynamics change drastically when it 
transitions between sticking and sliding contact or makes and breaks contact.

Our experiments show that despite of relying on depth images to extract
position and contact information, all our models perform similar 
to the analytical model applied on the ground truth state.
Given enough training data and evaluated inside of its training
domain, the pure neural network implementation performs best and even
outperforms the analytical model baseline significantly.
However, when it comes to generalization to new actions 
the hybrid approach is much more
accurate. Additionally, we find that the hybrid approach
needs significantly less training data than the neural network model
to arrive at a high prediction accuracy. 

\subsection{Contributions}

In this work, we make the following contributions: 
\begin{itemize}
\item We show how analytical dynamics models and neural networks can be 
combined and trained end-to-end to predict the effects of robot 
actions based on depth images or organized point clouds. 
\item We compare this hybrid approach to using a pure neural network for 
learning both, perception and prediction. Evaluations on a real world physical 
interaction task demonstrate improved data efficiency and
generalization when including the analytical 
model into the network over learning everything from scratch.
\item We show how the hybrid approach can be further extended by combining
the analytical model with a learned error-correction term to better compensate 
for possible inaccuracies of the analytical model
\item For training and evaluation, we augmented an existing dataset of planar pushing 
with depth and RGB images and additional contact information. The code for 
this is available online.
\end{itemize}

\subsection{Outline}

This paper is structured as follows: We begin with a review of related work
in Section~\ref{sec:related-work}. In Section~\ref{sec:problem} we formally 
describe the problem we address, introduce an analytical 
model for planar pushing (\ref{sec:model}) and the dataset we used for our 
experiments (\ref{sec:data}).
Section~\ref{sec:networks} introduces and compares the different approaches 
for learning perception and prediction. The evaluation in Section~\ref{sec:results2d} 
compares data-efficiency (\ref{sec:results-exp1}) and generalization abilities  
(\ref{sec:results-exp2}, \ref{sec:results-exp3}, \ref{sec:results-exp4}) of
the different architectures. The perception task is kept simple for these 
experiments by using a top-down view of the scene. This changes in Section~\ref{sec:3d}
where we demonstrate that the hybrid approach also performs well in a less constrained
visual setup. Section~\ref{sec:discussion} finally summarizes our results and gives an 
outlook to future work.

\section{Related Work} \label{sec:related-work}

\subsection{Models for Pushing}

Analytical models of quasi-static planar 
pushing have been studied extensively in the past, starting 
with \citet{mason}. \citet{goyal} introduced the limit surface
to relate frictional forces with object motion, and much 
work has been done on different approximate representations
of it \citep{lee-cutkosky, howe-cutkosky}. 
In this work, we use a model by \citet{model}, which relies on 
an ellipsoidal approximation of the limit surface. 

More recently, there has also been a lot of work on data-driven
approaches to pushing \citep{zhou, belter2014, mericcli, kopicki2017, bauza}.
\citet{kopicki2017} describe a modular learner that outperforms a physics
engine for predicting the results of 3D quasi-static pushing even for 
generalizing to unseen actions and object shapes. This is achieved by 
providing the learner not only with the trajectory of the global object 
frame, but also with multiple local frames that describe contacts. The 
approach however requires knowledge of the object pose from an external 
tracking system and the learner does not place the contact-frames itself.
\citet{bauza} train a heteroscedastic Gaussian Process
that predicts not only the object movement under a certain push,
but also the expected variability of the outcome. The trained model
outperforms an analytical model \citep{model} given very few training
examples. It is however specifically trained for one object and 
generalization to different objects is not attempted.
Moreover, this work, too, assumes access to the ground truth state, 
including the contact point and the angle between the push and the object 
surface.

\subsection{Learning Dynamics Based on Raw Sensory Data}

Many recent approaches in reinforcement learning aim to solve the so 
called ``pixels to torque'' problem, where the network 
processes images to extract a representation of the state 
and then directly returns the required action to achieve a certain task 
\citep{rf1,rf2}. 
\citet{robotic_priors} argue that the state-representation learned by such 
methods can be improved by enforcing \textit{robotic priors} on the extracted 
state, that may include e.g.\ temporal coherence. This 
is an alternative way of including basic principles of physics in a learning 
approach, compared to what we propose here.
While policy learning requires understanding the effect of actions, 
the above methods do not acquire an explicit dynamics model. We are interested 
in learning such an explicit model, as it enables optimal action selection 
(potentially over a larger time horizon). The following papers share this aim.

\citet{poke} consider a learning approach for pushing objects. 
Their network takes as input the pushing action and a pair of
images: one before and one after a push. After encoding the images, 
two different network streams attempt to predict (i) the encoding of the
second image given the first and the action and (ii)
the action necessary to transition from the first to the second encoding. 
Simultaneously training for both tasks improves the
results on action prediction. The authors do not enforce any physical
models or robotic priors. As the learned models directly operate
on image encodings instead of physical quantities, we cannot compare
the accuracy of the forward prediction part (i) to our results. 

SE3-Nets \citep{se3} process organized (i.e.\ image shaped) 3D point clouds 
and an action to predict the next point cloud. For each object in the scene, 
the network predicts a 
segmentation mask and the parameters of an SE3 transform (linear velocity, 
rotation angle and axis). 
In newer work \citep{se3pose}, an intermediate step is added, that computes 
the 6D pose of each object, before predicting the transforms based on this 
more structured state representation. The output point cloud is obtained by 
transforming all input pixels according to the transform for the object they 
correspond to. The resulting predictions are very sharp and the network is 
shown to correctly segment the objects and determine which are affected by 
the action. An evaluation of the generalization to new 
objects or forces was however not performed.  

Our own architecture is inspired by this work. The pure neural 
network we use to compare to our hybrid approach can be seen as a simplified 
variant of SE3-Nets, that predicts SE2 transforms (see Sec.~\ref{sec:networks}). 
Since we define the loss directly on the predicted movement of the object, 
we omit predicting the next observation and the segmentation masks required for 
this. We also use a modified perception network, which relies mostly on a small 
image patch around the robot end-effector.

\citet{NIPS2016_6161} is similar to \citep{se3} and explores different 
possibilities of predicting the next frame of a sequence of actions and RGB 
images using recurrent neural networks.

Visual Interaction Networks \citep{visual_interaction} also take 
temporal information into account. A convolutional neural network encodes
consecutive images into a sequence of object states. Dynamics are 
predicted by a recurrent network that considers 
pairs of objects to predict the next state of each object. 

\subsection{Combining Analytical Models and Learning}

The idea of using analytical models in combination with learning 
has also been explored in previous work. \citet{differentiabe_pysics_engine} 
implemented a differentiable physics 
engine for rigid body dynamics in Theano and demonstrate how it can be used 
to train a neural network controller. In \citep{gp}, the authors 
significantly improve Gaussian Process learning of inverse dynamics by using 
an analytical model of robot dynamics with fixed parameters as the mean 
function or as feature transform inside the covariance function of the
GP. Both works however do not cover visual perception. 
Most recently, \citet{deanimation} used a graphics and physics engine 
to learn to extract object-based state representations in an 
unsupervised way: Given a sequence of images, a network learns to 
produce a state representation that is predicted forward in time using the 
physics engine. The graphics engine is used to render the
predicted state and its output is compared to the next image as training 
signal. 
In contrast to the aforementioned work, we not only combine learning and
analytical models, but also evaluate the advantages and limitations of this 
approach.  
Finally, \citep{eql} present an interesting approach to learning functions 
by training a neural network to combine a number of mathematical base 
operations (like multiplication, division, sine and cosine). This enables 
their ``Equation Learner'' to learn functions which generalize beyond 
the domain of the training data, just like traditional analytical models. 
Training these networks is however 
challenging and involves training many different models and choosing the
best in an additional model selection step.

\section{Problem Statement}\label{sec:problem}
Our aim is to analyse the benefits of combining neural networks 
with analytical models. We therefore compare this hybrid approach to models that 
exclusively rely on either approach. As a test bed, we use planar
pushing, for which a well-known analytical model and a real-world dataset are 
available. 

We consider the following problem: The input consists of a depth image $\mathbf{D}^t$ 
of a tabletop scene with one object and the pusher at time $t$, the starting position 
$\mathbf{p}^t$ of the pusher and its movement between this and the next timestep  
$\mathbf{u}^{t} = \mathbf{p}^{t+1} - \mathbf{p}^t$.
With this information, the models need to predict the object position $\mathbf{o}^t$ before
the push is applied and its movement $\mathbf{v_o}^{t} = \mathbf{o}^{t+1} - \mathbf{o}^t$ due 
to the push.

This can be divided into two subproblems: 
\paragraph{Perception} 
Extract a suitable state representation of the scene at time $t$ (before the push)
$\mathbf{x}^t$ from the input image. The form of $\mathbf{x}^t$ depends on the following
prediction model, we only require that $\mathbf{x}^t$ contains the object position 
$\mathbf{o}^t$. 
\begin{equation*}f_{perception}(\mathbf{D}^t) = \mathbf{x}^t
\end{equation*}

\paragraph{Prediction} 
Given the state representation $\mathbf{x}^t$, the position $\mathbf{p}^t$ of the pusher 
and its movement $\mathbf{u}^{t:t+1}$, predict how the object will move: 
\begin{equation*}f_{prediction}(\mathbf{x}^t, \mathbf{p}^t, \mathbf{u}^{t:t+1}) = \mathbf{v_o}^{t:t+1}\end{equation*}

In the following sections, we will introduce an analytical model for computing 
$f_{prediction}$ and the dataset of real robot pushes that we use for training 
and evaluation.

\subsection{An Analytical Model of Planar Pushing}\label{sec:model}

\begin{figure*}
\begin{subfigure}{0.16\textwidth}
\includegraphics[width=\textwidth]{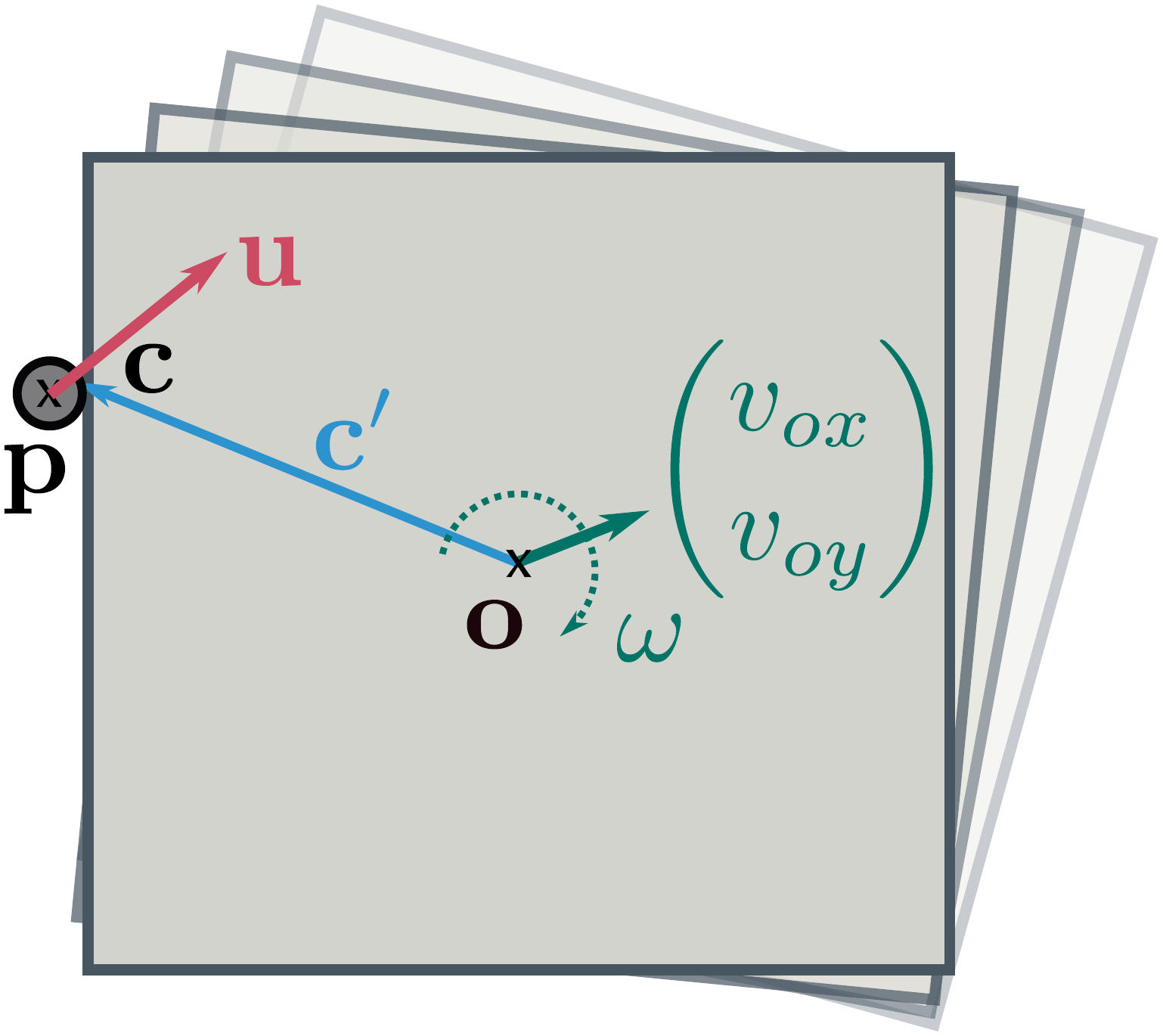}
\end{subfigure}
\begin{subfigure}{0.34\textwidth}
\footnotesize
\begin{tabular}{c p{4.25cm}}
$\mathbf{o}$ & position of the object \\
$\mathbf{v_o}$ & linear and angular object velocity \\
$\mathbf{v_p}$ & linear velocity at the contact point - \textit{effective push velocity} \\
$\mathbf{p}$ & position of the pusher \\
$\mathbf{u}$ & linear pusher velocity - \textit{action} \\
$\mathbf{c}$ & contact point (global) \\
$\mathbf{c'}$ & contact point relative to $\mathbf{o}$ \\
$\mathbf{n}$ & surface normal at $\mathbf{c}$ \\
$l$ & ratio between maximal torsional and linear friction force   \\
$\mu$ & friction coefficient pusher-object \\
\end{tabular}
\end{subfigure}
\begin{subfigure}{0.19\textwidth}
\includegraphics[width=\textwidth]{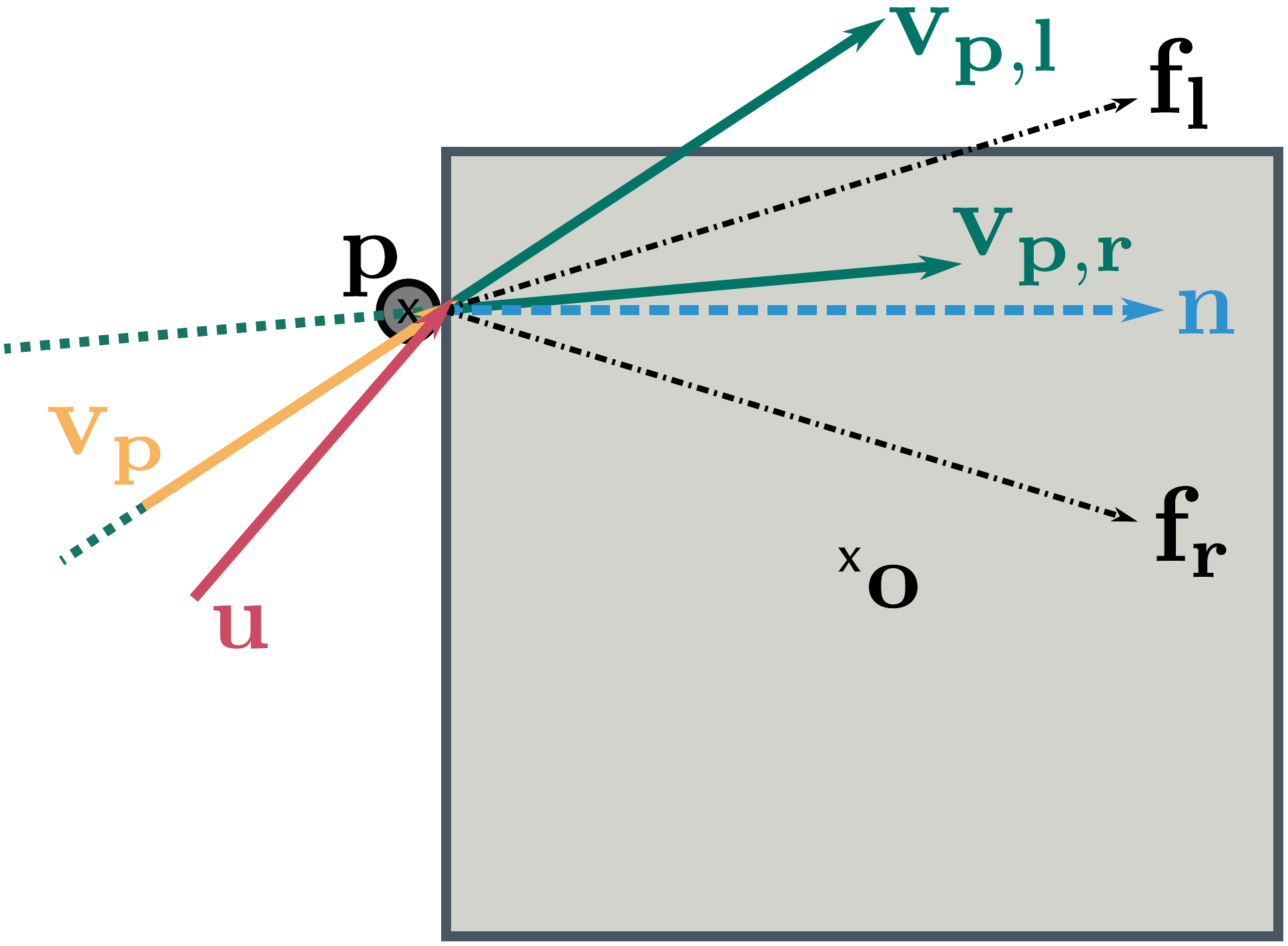}
\end{subfigure}
\begin{subfigure}{0.25\textwidth}
\footnotesize
\begin{tabular}{c p{3.75cm}}
$\mathbf{f}_b $  & left or right boundary force of the friction cone \\
$m_b$ & torques corresponding to the boundary forces \\
$\mathbf{v_{o,b}}$ & object velocities resulting from boundary forces \\
$\mathbf{v_{p,b}}$ & effective push velocities corresponding to the boundary forces \\
$ b = {l,r}$ & placeholder for left or right boundary \\
$s$ & contact indicator, $s \in [0, 1]$ \\
$\mathbf{k}$ & rotation axis
\end{tabular}
\end{subfigure}
\caption{Overview and illustration of the terminology for
  pushing.} \label{fig-symbols}
\end{figure*}

We use the analytical model of quasi-static planar pushing that was 
devised by \citet{model}. It predicts the object movement $\mathbf{v_o}$
given the pusher velocity $\mathbf{u}$, the contact point $\mathbf{c}$
and associated surface normal $\mathbf{n}$ as well as two 
friction-related parameters $l$ and
$\mu$. The model is illustrated in Figure~\ref{fig-symbols}, which
also contains a list of symbols. Note that this model is still
approximate and far from perfectly modelling the stochastic process of
planar pushing~\citep{data}.

Predicting the effect of a push with this model has two stages: First,
it determines whether the push is stable (``sticking
contact'') or whether the pusher will slide along the object
(``sliding contact''). In the first case, the velocity of the object at 
the contact point will be the same as the velocity of the pusher. In the 
sliding case however, the pusher movement can be almost orthogonal to the 
resulting motion at the contact point. We call the motion at the contact 
point ``effective push velocity'' $\mathbf{v_p}$. It is the output of the first 
stage. Given $\mathbf{v_p}$ and the contact point, the second stage then 
predicts the resulting translation and rotation of the object's centre of mass. 

\subsubsection*{Stage 1: Determining the Contact Type and Computing $\mathbf{v}_p$}
To determine the contact type (slipping or sticking), we have to 
find the left and right boundary forces $\mathbf{f_{l}}$,
$\mathbf{f_{r}}$ of the friction cone (i.e.\ the forces for which the pusher 
will just not start sliding along the object) and the corresponding torques 
$m_l$, $m_r$. The opening angle $\alpha$ of the friction cone is defined by the 
friction coefficient $\mu$ between pusher and object. The forces and torques are then computed by  
\begin{align}
\alpha &= \arctan (\mu)  \\
\mathbf{f}_{l} &= \mathbf{R(- \alpha)} \mathbf{n}  & \mathbf{f}_{r} &= \mathbf{R(\alpha)} \mathbf{n}  \\
m_{l} &= c'_x f_{ly} - c'_y f_{lx} & m_{r} &= c'_x f_{ry} - c'_y f_{rx}
\end{align} where  $\mathbf{R(\alpha)}$ denotes a rotation matrix
given $\alpha$ and $\mathbf{c'} = \mathbf{c} - \mathbf{o}$ is the  
contact point relative to the object's centre of mass.

To relate the forces to object velocities, \citet{model} use an
ellipsoidal approximation to the limit surface. To simplify notation,
we use subscript $b$ to refer to quantities associated with either the
left $l$ or right $r$ boundary forces.
$\mathbf{v_{o,b}}$ and $\omega_{o,b}$ denote linear and angular object
velocity, respectively. $\mathbf{v_{p,b}}$ 
are the push velocities that would create the boundary forces. They span 
the so called ''motion cone''.
\begin{align}
\mathbf{v_{o,b}} &=  \frac{\omega_{o,b} l^2 }{m_{b}} \mathbf{f_{b}} \\
\mathbf{v_{p,b}}  &=\omega_{o,b} (\frac{l^2 } {m_{b}} \mathbf{f_{b}}  +  \mathbf{k} \times \mathbf{c'} ) 
\end{align}
$\omega_{o,b}$ acts as a scaling factor. Since we are only interested
in the direction of $\mathbf{v_{p,b}}$ and not in its magnitude, we
set  
$\omega_{o,b} = m_b$:
\begin{align}
\mathbf{v_{p,b}} &= l^2 \mathbf{f_{b}} + m_{b}  \mathbf{k} \times \mathbf{c'}
\end{align}

To compute the effective push velocity $\mathbf{v_p}$, we need to determine 
the contact case: If the push velocity lies outside of the motion cone, the 
contact will slip. The resulting effective push velocity then acts in the 
direction of the boundary velocity $\mathbf{v_{p,b}}$ which is closer to the 
push direction: 
\begin{align}
\mathbf{v_p}  &= \frac{\mathbf{u} \cdot \mathbf{n}}{\mathbf{v_{p,b}} \cdot 
\mathbf{n}} \mathbf{v_{p,b}} 
\end{align} 
Otherwise contact is sticking and we can use the pusher velocity as 
effective push velocity $\mathbf{v_p} = \mathbf{u}$.
When the norm of $\mathbf{n}$ is 
zero (due to e.g.\ a wrong prediction of the perception neural
network), we set the output $\mathbf{v_{p,b}}$ to zero. 

The object will of course only move if the pusher is in contact with the 
object. To use the model also in cases where no force acts on the object, we 
introduce the contact indicator variable $s$. It takes values between zero 
and one and is multiplied with $\mathbf{v_p}$ to switch off responses when 
there is no contact. 
\begin{align*}
\mathbf{v_p}  &= s \hspace{1pt} \mathbf{v_p}  
\end{align*}

We allow $s$ to be continuous instead of binary to give the model a chance to 
react to the pusher making or breaking contact during the interaction.

\subsubsection*{Stage 2: Using $\mathbf{v_p}$ to Predict the Object Motion} 
\label{sec:model-pt2}
Given the effective push velocity $\mathbf{v_p}$ and the contact point 
$\mathbf{c}'$ relative to the object centre of mass, we can compute
the  linear and angular velocity $\mathbf{v_o} = [v_{ox}, v_{oy},  
\omega]$ of the object.

\begin{align}
v_{ox} &=  \frac{(l^2 + c_x^{\prime 2}) v_{px} + c'_x c'_y v_{py}}{l^2 + c_x^{\prime 2} + c_y^{\prime 2}} \\
v_{oy} &=  \frac{(l^2 + c_y^{\prime 2}) v_{py} + c'_x c'_y v_{px}}{l^2 + c_x^{\prime 2} + c_y^{\prime 2}} \\
\omega &= \frac{c'_x v_{oy} - c'_y v_{ox}}{l^2} 
\end{align}

\subsubsection*{Discussion of Underlying Assumptions}
The analytical model is built on three simplifying assumptions: (i) quasi-
static pushing, i.e.\ the force applied to the object is big enough 
to move the object, but not to accelerate it (ii) the pressure 
distribution of the object on the surface is uniform and the 
limit-surface of frictional forces can be approximated by an ellipsoid 
(iii) the friction coefficient between surface and object is constant. 

The analysis performed by \citet{data} shows that assumption (ii) and 
(iii) are violated frequently by real world data. Assumption (i) holds 
for push velocities below 50\,$\frac{mm}{s}$. In addition, the contact 
situation may change during pushing (as the pusher may slide along the 
object and even lose contact), such that the model predictions become 
increasingly inaccurate the longer ahead it needs to predict in one step.

\subsection{Data} \label{sec:data}

We use the MIT Push Dataset~\citep{data} for our experiments.
It contains object pose and force recordings (not used here)
from real robot experiments, 
where eleven different planar objects are pushed on four different surfaces. 
For each object-surface combination, the dataset contains about 
6000 pushes that vary in the manipulator (``pusher'') 
velocity and acceleration, the point on the object where the pusher 
makes contact and the angle between the object surface and the push  
direction. Pushes are 5\,cm long and data was recorded at 250\,Hz.

\begin{figure}
\centering
\includegraphics[width=\columnwidth]{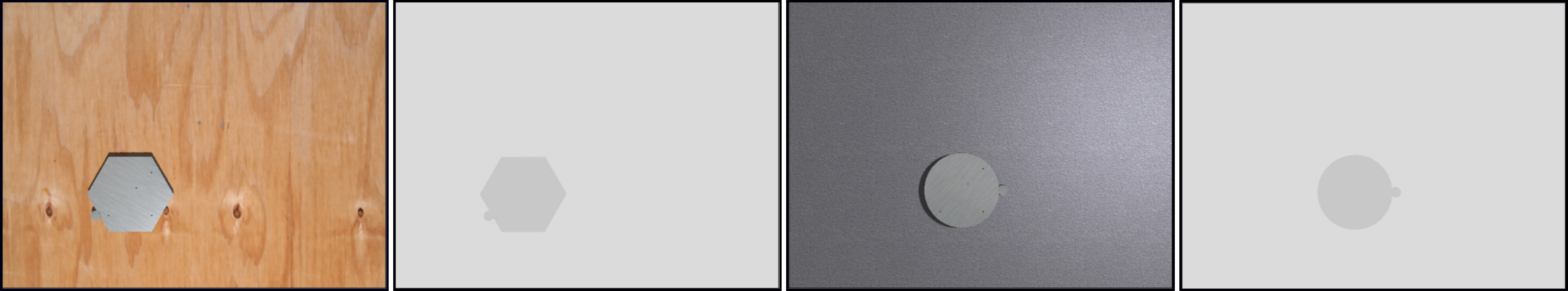}
\caption{\label{fig-data-b} Rendered RGB and depth 
images on two of the four surfaces in the MIT dataset,
\textit{plywood} and \textit{abs}.} 
\end{figure}

As this dataset does not contain RGB or depth images, we render them
using OpenGL and the mesh-data supplied with the dataset. In this work, 
we only use the depth images, RGB will be considered 
in future work.  A rendered scene consists of a flat surface with one of four 
textures (representing the four surface materials),
on which one of the objects is placed. The pusher is represented by a
vertical cylinder with no arm attached. Figures~\ref{fig-data-a} and \ref{fig-data-b} 
show the different objects and example images. 
We also annotated the dataset with all information necessary to apply
the analytical model to use it as a baseline. The code for annotation
and rendering is available \href{https://github.com/mcubelab/pdproc}{here}. 

For each experiment, we construct datasets for training and testing from a subset
of the Push Dataset. As the analytical model does not take acceleration 
of the pusher into account, we only use push variants with zero 
pusher acceleration. We however do evaluate on data with
high pusher velocities, that break the quasi-static assumption made in
the analytical model (in Sec.~\ref{sec:results-exp3}). 
One data point in our datasets consists of a depth image showing the 
scene before the push is applied, the object position before and after the
push and the initial position and movement of the pusher. The prediction horizon
is 0.5\,seconds in all datasets~\endnote{We also evaluated two
different prediction horizons but
found no significant effect on the performance.}. More information about the
specific datasets for each experiment can be found in the corresponding  sections.

We use data from multiple randomly chosen timesteps of each sequence in the 
Push Dataset. Some of the examples thus contain shorter push-motions
than others, as the pusher starts moving with some delay or  
ends its movement during the 0.5\,seconds time-window. 
To achieve more visual variance and to balance the number of examples per
object type, we sample a number of transforms of the scene relative to the 
camera for each push. Finally, about a third of our dataset consists of 
examples where we moved the pusher away from the object, such that it 
is not affected by the push movement.

\section{Combining Neural Networks and Analytical Models} \label{sec:networks}

We now introduce the neural network variants that we will analyse in 
the following section. All architectures share the same first network stage 
that processes raw depth images and outputs a lower-dimensional encoding and the
object position. Given this output, the pushing
action (movement $\mathbf{u}$ and position
$\mathbf{p}$) of the pusher, and the friction parameters $\mu$ and $l$\endnote{We provide 
these inputs as friction related information cannot be 
obtained from single images. Estimation from sequences is considered
future work.}, the second part of these networks predicts the  
linear and angular velocity $\mathbf{v_o}$ of the object. 
This predictive part differs between the network variants. While three of 
them (\textit{simple, full, error}) use variants of the analytical dynamics model 
established in Sec.~\ref{sec:model}, variant \textit{neural} has to learn 
the dynamics with a neural network. The prediction part 
has about 1.8\,million trainable parameters for all variants except for \textit{error},
which has 2.7\,million parameters.

We implement all our networks as well as the analytical model in
tensorflow~\citep{tensorflow}, which allows us to propagate gradients
through the analytical models just like any other layer.

\subsection{Perception} \label{sec:methods-perc}
\begin{figure}
\centering
\includegraphics[width=\columnwidth]{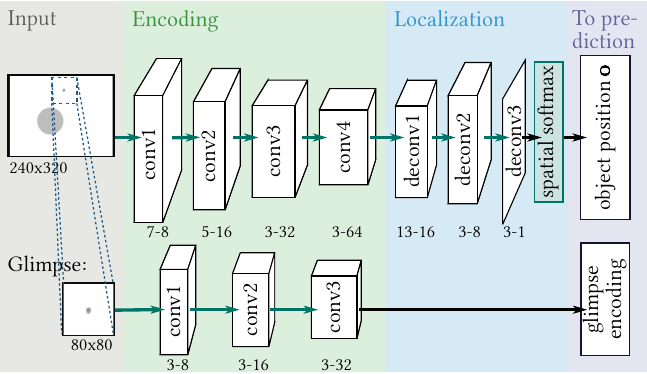}
\caption{Perception part for all network variants. White boxes represent tensors, 
green arrows and boxes indicate network layers, whereas black arrows represent dataflow 
without processing. For green arrows, the type of layer (convolution or deconvolution) 
is denoted in the name of their output tensors. The numbers below the output tensors denote 
the kernel size and the number of output channels for each layer.\\
The output of this module, glimpse encoding and the estimated object position $\mathbf{o}$,
serves as input for the prediction network depicted in Figure~\ref{fig-pred}.
For training, gradient information is backpropagated through the prediction
to the perception network.
}\label{fig-perc}
\end{figure}

The architecture of the network part that processes the image is depicted in 
Fig.~\ref{fig-perc}. 
We assume that the robot knows the position of its end-effector, which allows us 
to extract a small ($80 \times 80$  pixel) image patch
(``glimpse'') around the tip of the pusher. If the pusher is close enough to
the object to make contact, the important information for predicting the effect 
of the push - like the contact point and the normal to the object surface - can 
be estimated from this smaller image. It thus serves as an attention-mechanism 
to focus the computations on the most relevant part of the image. 
Only the position of the object needs to be estimated from the full image. 
The state representation that our perception model extracts thus contains the
estimated object position and an encoding of the information represented
in the glimpse.
%Taken together, this is all the information necessary to
%predict the object movement.

To obtain the glimpse encoding, we process the glimpse with three convolutional 
layers with ReLU non-linearity, each followed by max-pooling and batch
normalization~\citep{batch_norm}. For estimating the object position, 
the full image is processed with a sequence of four convolutional and three 
deconvolution layers. The output of the last deconvolution has the same size as the
image input and only has one channel that resembles an object segmentation map. 
We use spatial softmax \citep{rf2} to calculate the pixel location of the 
segmented object centre.  

Initial experiments showed that not using the glimpse strongly decreased 
performance for all networks. We also found that using both, the glimpse and an 
encoding of the full image, for estimating all physical parameters was 
disadvantageous: Using the full image increases the number of trainable 
parameters in the prediction network but adds no information that is not 
already contained in the glimpse.

% moved here to avoid the weird pdflink error...
\begin{figure}
\centering
\includegraphics[width=\columnwidth]{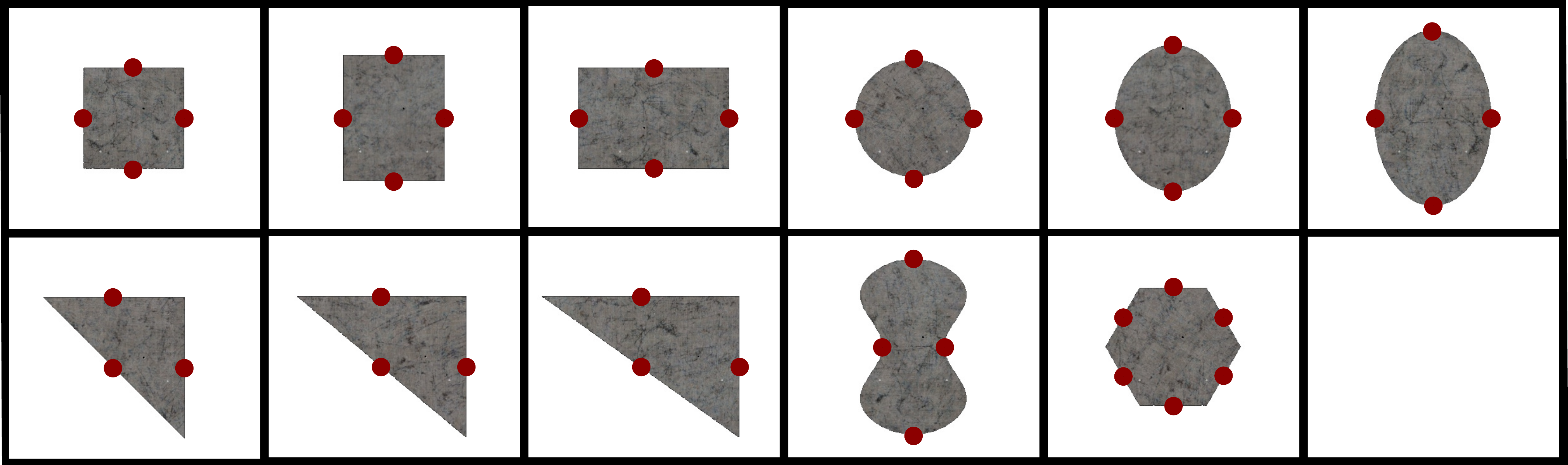} 
\caption{\label{fig-data-a}Rendered objects of the Push
  Dataset~\citep{data}: \textit{rect1-3, ellip1-3, tri1-3, butter, hex}. 
  Red dots indicate the subset of contact points we use
  to collect a test set with held-out pushes for Experiment~\ref{sec:results-exp2}.}
\end{figure}

\subsection{Prediction}

\paragraph{Neural Network only (\textit{neural})}

\begin{figure}
\centering
\includegraphics[width=\columnwidth]{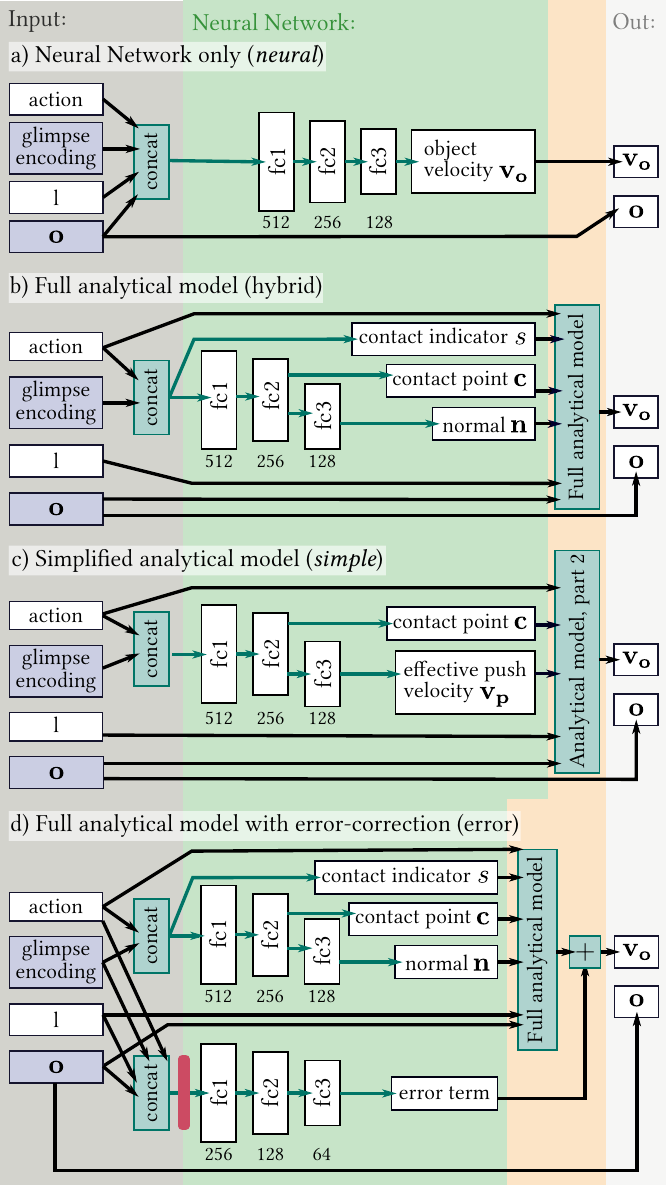}
\caption{Prediction parts of the four network variants 
\textit{neural}, \textit{hybrid}, \textit{simple} and \textit{error}. 
White and purple boxes represent tensors, where the purple color indicates
tensors that are computed by the perception part shown in Figure~\ref{fig-perc}.
During training, the gradient information is backpropagated through these 
tensors to the perception part.
\\
Green arrows and boxes indicate network layers, whereas black arrows represent dataflow 
without processing. In this network, all green arrows represent fully-connected layers
and the numbers beneath their output tensors (fc) denote the number of output channels.
The red bar in architecture (d) indicates that no gradients are propagated to the inputs
of this layer. 
}\label{fig-pred}
\end{figure}

Figure~\ref{fig-pred} a) shows the prediction part of the 
variant \textit{neural}, which uses a neural network to learn the 
dynamics of pushing. The input to this part is a concatenation of the
output from perception with the action and friction parameter $l$.
The network processes this input with 
three fully-connected layers before predicting the object velocity 
$\mathbf{v_o}$. All intermediate fully-connected layers use ReLU
non-linearities. The output layers do not apply a non-linearity.  

\paragraph{Full analytical model (\textit{hybrid})}
This variant uses the complete analytical model as described in Section 
\ref{sec:model}. Several fully-connected layers  extract the 
necessary input values from the glimpse encoding 
and the action, as shown in Figure~\ref{fig-pred} b). These are 
the contact point $\mathbf{c}$, the surface normal $\mathbf{n}$
and the contact indicator $s$. For predicting $s$, we use a
sigmoidal non-linearity to limit the  predicted values to 
$\left[0, 1\right]$.

\paragraph{Simplified analytical model (\textit{simple})}
\textit{Simple} (Figure~\ref{fig-pred} c) only uses the
second stage of the analytical model. As for \textit{hybrid}, a neural network 
extracts the model inputs (effective push velocity $\mathbf{v_p}$, 
contact point $\mathbf{c}$) from the encoded glimpse and the action. 

We use this variant as a middle ground between the two other options: 
It still contains the main mechanics of how an effective push at the 
contact point moves the object, but leaves it to the neural network to 
deduce  the effective push velocity from the scene and the action.
This gives the model more freedom to correct for possible shortcomings 
of the analytical model. We expect these to manifest mostly in the first 
stage of the model, as small errors can have a big effect there
when they influence whether a contact is estimated as sticking or slipping.
Since the second stage of the analytical model does not specify how the input 
action relates to the object movement, \textit{simple} also allows 
us to evaluate the importance of this particular aspect of the analytical 
model.

\paragraph{Full analytical model + error term (\textit{error})}
One concern when using a predefined analytical model is that the trained
network cannot improve over the performance of the analytical model. If the analytical
model is inaccurate, the \textit{hybrid} architecture can only
compensate to some degree by manipulating the input values of the model, i.e.\
by predicting ``incorrect'' values for the components of the state representation. 
This limits its ability to compensate for model errors as it might not be possible 
to account for all types of errors in this way.

As an alternative, we propose to learn an error-correction term which is added
to the output prediction of the analytical model. The error-term is thus not
constrained by the model and should be able to compensate for a broader class
of model errors.

Figure~\ref{fig-pred} d) shows the architecture. As input for predicting the  
error-term, we use the same values that \textit{neural} receives for predicting
the object velocity, i.e.\ the glimpse encoding, the action, the predicted object
position and the friction parameter. Note that we do not propagate gradients to 
the inputs of the error-prediction module. The intuition behind this is that
we do not want the error-prediction to interfere with the prediction of the 
inputs for the analytical model. We evaluate the effect of this architectural 
decision in Section~\ref{sec:results-exp5}. A second variant that we compare to in
this section aims to improve the generalizability of the error-prediction to faster push 
movements. This is achieved by normalizing the input action to unit length before
feeding it into the error-prediction module.

\subsection{Training}

All our architectures are trained end-to-end, i.e.\ the loss is propagated through the 
prediction to the perception part of the networks. 
The loss $L$ penalizes the Euclidean distance between the predicted and the 
real object position in the input image (\textit{pos}), the Euclidean error 
of the predicted object translation (\textit{trans}), the error in the 
magnitude of translation (\textit{mag}) and in 
angular movement (\textit{rot}) in degree (instead of radian, 
to ensure that all components of the loss have the same order of
magnitude). We use weight decay with $\lambda = 0.001$.

Let $\mathbf{\hat{v}_o}$ and  $\mathbf{\hat{o}}$ denote the predicted and  
$\mathbf{v_o}$, $\mathbf{o}$ the real object movement and position. 
$\mathbf{w}$ are the network weights 
and $\mathbf{\nu_{o}} = [v_{ox}, v_{oy}]$  denotes linear object velocity.
\begin{gather*}
L(\mathbf{\hat{v}_o}, \mathbf{\hat{o}}, \mathbf{v_o}
, \mathbf{o}) = trans + mag + rot +  pos + \lambda \sum\nolimits_{\mathbf{w}} \parallel \mathbf{w} \parallel \\
\begin{aligned}
    trans &= \left\| \mathbf{\hat{\nu}_{o}} - \mathbf{\nu_{o}} \right\| &
mag &= \left|  \left\| \mathbf{\hat{\nu}_{o}} \right\| - \left\| \mathbf{\nu_{o}}  \right| \right\| \\
rot &= \tfrac{180}{\pi}  | \omega - \hat{\omega} | &
pos &= \parallel \mathbf{o} - \mathbf{\hat{o}} \parallel
\end{aligned}
\end{gather*}

When using the variant \textit{hybrid}, a major challenge is the contact 
indicator $s$: In the beginning of training, the direction of the predicted 
object movement is mostly wrong. $s$ therefore receives a strong negative 
gradient, causing it to decrease quickly. 
Since the predicted motion is effectively multiplied by $s$, a low $s$ 
results in the other parts of the network receiving small gradients and 
thus greatly slows down training. We therefore add the error in 
the magnitude of the predicted velocity to the loss to prevent $s$ from 
decreasing too far in the early training phase.

We use Adam optimizer~\citep{adam} with a learning rate of 
$0.0001$ and a batch-size of 32 for 75,000 steps. 

\section{Evaluating Generalization}\label{sec:results2d}

In this section, we test our hypothesis that using an analytical model for 
prediction together with a neural network for perception improves data 
efficiency and leads to better generalization than using neural networks for 
both, perception and prediction. 
We evaluate how the performance of the networks depends on the amount
of training data (Experiment \ref{sec:results-exp1}) and how well they generalize to (i)
pushes with new pushing angles and contact points (Experiment \ref{sec:results-exp2}), 
(ii) new push velocities (Experiment \ref{sec:results-exp3}) and (iii) unseen objects
(Experiment \ref{sec:results-exp4}).

For the experiments in this section, we use a top-down view of the
scene, such that the object can only move in the image plane and the
$z$-coordinate of all scene components remains constant. This
simplifies the application of the analytical model by removing the
need for an additional transform between the camera and
the table. It also simplifies the perception task and allows us to focus this
evaluation on the comparison of the hybrid and the purely neural approach. 
In Section~\ref{sec:3d} we will show how to extend the proposed model 
to work on more difficult camera settings. 

\subsection{Baselines}

We use three baselines in our experiments. All of them use the ground 
truth input values of the analytical model (action, object position, contact 
point, surface normal, contact indicator and friction coefficients) instead of 
depth images. They thus do not solve the full problem of predicting object
movement from raw sensory input. Instead, they address the easier problem of prediction 
given perfect state information. Accordingly, the baselines only output the 
object velocity, but not its initial position in the scene. 

If the pusher makes contact with the object during the push, but is not in contact 
initially, we use the contact 
point and normal from when contact is first made and shorten the action 
accordingly. Note that this gives the baseline models an additional advantage over 
architectures that have to infer such input values from raw sensory data.

The first baseline is just the average translation and rotation over the 
dataset. This is equal to the error when always predicting zero movement, and 
we therefore name it \textit{zero}. 
The second,\textit{physics}, is the full analytical model
evaluated on the ground truth input values. 
The third baseline, called \textit{neural dyn} is a neural network that 
has the same three-layer architecture as the prediction module of \textit{neural} 
(see Figure~\ref{fig-pred} a) for details). The difference
between \textit{neural} and \textit{neural dyn} is their input:
While \textit{neural} receives the glimpse encoding and object position from the 
perception network as input, 
\textit{neural dyn} gets the ground truth physical state representation that is
also used in the analytical model. This allows us to evaluate whether \textit{neural} 
benefits from being able to learn its own state representation (the glimpse encoding)
end-to-end through the prediction part.

% In addition to the networks described in Section~\ref{sec:networks}, 
% we also train a neural networkto predict the object 
%movement given the same state representation as used by the analytical model. 
%It uses the same architecture of three fully-connected layers for prediction 
%as \textit{neural} (see Figure~\ref{fig-pred} a) for details), with the difference
%that \textit{neural} receives input from the perception network while \textit{neural dyn}
%gets the ground truth physical state representation. 
%This baseline allows us to evaluate whether \textit{neural} benefits from being able
%to choose its own state representation. 

\subsection{Metrics}

For evaluation, we compute the average Euclidean distance between the 
predicted and the ground truth object translation (\textit{trans}) and 
position (\textit{pos}) in millimeters as well as the average error on object 
rotation (\textit{rot}) in degree. As our datasets differ in the overall 
object movement, we report errors on translation 
and rotation normalized by the average motion in the corresponding dataset
given by the error of the baseline \textit{zero}.

\subsection{Data Efficiency}\label{sec:results-exp1}

The first hypothesis we test is that combining the analytical model
with a neural network for perception reduces the required training 
data as compared to a pure neural network. 

\subsubsection*{Data}

We use a dataset that contains all objects from the MIT Push dataset and 
all pushes with velocity $20\frac{mm}{s}$ and split it randomly into training
and test set. This results in about 190k training examples and about 38k 
examples for testing. To evaluate how the networks' performance develops with
the amount of training data, we train the models on different subsets of the 
training split with sizes from 2500 to the full 190k. We always evaluate on the
full test split. To reduce the influence of dataset composition especially on the 
small datasets, we average results over multiple different
datasets with the same size.

\subsubsection*{Results}

\begin{figure*}
\centering
\includegraphics[width=\textwidth]{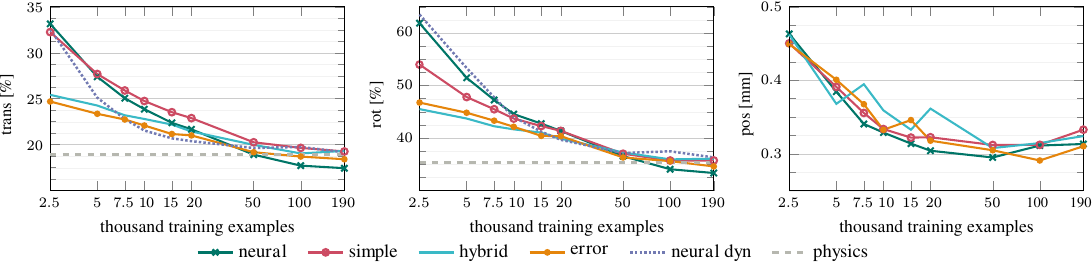}
\caption{Prediction errors versus training set size (x-axis in logarithmic    
  scale). Errors on translation and
  rotation are given as percentage of the average
  movement in the test set. The model-based architecture \textit{hybrid} performs much
  better than the other networks when training data is
  sparse.}\label{fig-datasize}
\end{figure*}

Figure~\ref{fig-datasize} shows how the errors in predicted translation, 
rotation and object position develop with more training data and Table~\ref{tab-exp1} 
contains numeric values for training on the biggest and 
smallest training split. As expected,
the combined approach of neural network and analytical model (\textit{hybrid}
and \textit{error})
already performs very well on the smallest dataset (2500 examples) and beats 
the other models including the \textit{neural dyn} baseline, which uses the ground 
truth state representation, by a large margin.
It takes more than 20k training examples for the other models to reach
the performance of \textit{hybrid}, where predicting rotation seems to be harder 
to learn than translation.

Despite of having to rely on raw depth images instead of the ground truth 
state representation, all models perform at least close to the 
\textit{physics} baseline when using the full training set. However, only the 
pure neural network and the hybrid model with error-correction are able 
to improve on the baseline. This shows that the analytical model limits \textit{hybrid} 
in fitting the training data perfectly, since the model itself is not perfect and does not 
allow for overfitting to noise in the training data. \textit{Neural} and \textit{error}
have more freedom for fitting the training distribution, which however also 
increases the risk of overfitting. 

Combining the learned error-correction with the fixed analytical model is especially helpful
for predicting the translation of the object. To also improve the prediction of rotations, 
the model needs more than 20k training examples, which is similar to \textit{neural}.
While \textit{neural} makes a larger improvement on the full dataset, \textit{error}
combines the comparably good performance of \textit{hybrid} on few training examples 
with the ability to improve on the model given enough data. 

The variant \textit{simple}, which uses only the second part of the analytical model,
also combines learning and a fixed model for predicting the dynamics. But in contrast to
\textit{error}, this variant seems to combine the disadvantages of both 
approaches: It needs much more training data than \textit{hybrid} but is still limited
by the performance of the analytical model and gets quickly outperformed by the pure 
neural network when more data is available.

The comparison of \textit{neural} and the baseline \textit{neural dyn} shows 
that despite of having access to the ground truth data, \textit{neural dyn} 
actually performs worse than \textit{neural} on the full dataset. 
This seems to agree with the theory of \citet{poke}, that training perception 
and  prediction end-to-end and letting the network
chose its own state representation instead of forcing it to use
a predefined state may be beneficial for neural learning.

\begin{table}
\scriptsize
\caption{Error in predicted translation (\textit{trans}) and 
rotation (\textit{rot}) as percentage of the average movement given by \textit{zero} (standard errors 
in brackets). \textit{pos} denotes the error in predicted object position.
Values shown are for training on the full training set (190k examples) and on a 
2500 examples subset. 
\label{tab-exp1}} 
\begin{tabular}{@{}c r l l l} 
\toprule
&  & trans  & rot   & pos [mm] \\
\midrule
\multirow{5}*{\rotatebox{90}{2.5k}}  
& \textit{neural} & $ 33.6\: (0.18)$\,\% & $ 62.54\: (0.42)$\,\% & $ 0.46\: (0.002)$ \\
& \textit{simple} & $ 32.3\: (0.19)$\,\% & $ 53.6 \: (0.37)$\,\%  & $\mathbf{0.44 \: (0.002)}$ \\
& \textit{hybrid} & $ 25.4\: (0.17)$\,\% & $ \mathbf{45.5\: (0.36)}$\,\% & $ 0.46\: (0.002)$  \\
& \textit{error} & $ \mathbf{24.7\: (0.16)}$\,\% & $ 46.8\: (0.36)$\,\% & $ 0.45\: (0.002)$  \\
& \textit{neural dyn}& $ 32.6 \: (0.19)$\,\% & $ 63.5 \: (0.46)$\,\% & -\\
\midrule
\multirow{5}*{\rotatebox{90}{190k}} 
& \textit{neural} & $\mathbf{17.4 \: (0.12)}$\,\% & $\mathbf{33.4 \: (0.28)}$\,\% & $\mathbf{0.31 \: (0.002)}$ \\
& \textit{simple} & $19.3 \: (0.13)$\,\% & $35.7 \: (0.3)$\,\% & $0.33 \: (0.002)$ \\
& \textit{hybrid}  & $19.3 \: (0.13)$\,\% & $36.1 \: (0.3)$\,\% & $0.32 \: (0.002)$ \\
& \textit{error}  & $ 18.4 \: (0.12)$\,\% & $ 34.6\: (0.29)$\,\% & $ \mathbf{0.31\: (0.002)}$  \\
& \textit{neural dyn} & $19.2 \: (0.12)$\,\% & $36.3 \: (0.29)$\,\% & - \\
\midrule
& \textit{physics}  &  $18.95 \: (0.13)$\,\% & $35.4 \: (0.3)$\,\% & -\\
& \textit{zero} & $2.95 \: (0.02)$\,mm & $1.9 \: (0.01)\, ^{\circ}$ & - \\
\bottomrule
\end{tabular}
\end{table}

Finally, we evaluate how accurate the predicted input values to the
analytical model are for \textit{simple}, \textit{hybrid} 
and \textit{error}.
If the analytical model was perfect, we would expect the predicted values
to be very close to the real physical state. Higher errors could thus indicate that the models
learn to compensate for inaccuracies of the analytical model.

\begin{table}
\scriptsize
\caption{Errors of the predicted input values for the analytical model: 
\textit{Hybrid} and \textit{error} predict the contact point 
$\mathbf{c}$, then normal $\mathbf{n}$ and the contact indicator $s$ accurately. 
\textit{Simple} only predicts the contact point and the effective push velocity $\mathbf{v_p}$, 
which both deviate notably from their ground truth values.
Values shown are for training on the full training set (190k examples). 
\label{tab-exp1-state}} 
\begin{tabular}{r l l l l} 
\toprule
  & $\mathbf{c}$ [mm] & $\mathbf{n}$ [$^{\circ}$] & $s$ & $\mathbf{v_p}$ [$^{\circ}$] \\
\midrule
\textit{simple} & $22.4 \: (0.027)$ & -                            & -                 & $18.1	 \: (0.111)$ \\
\textit{hybrid} & $4.4 \: (0.01)$   & $3.6 \: (0.024)$ & $0.08 \: (0.001)$ & - \\
\textit{error}  & $4.8 \: (0.011)$  & $2.5 \: (0.02)$  & $0.08 \: (0.001)$ & - \\
\bottomrule
\end{tabular}
\end{table}

As can be seen in Table~\ref{tab-exp1-state}, both \textit{hybrid} and \textit{error} 
make fairly accurate predictions for the object state, with contact point errors around
5\,mm and less than 5$^{\circ}$ angle between the predicted and correct normal. The contact
point indicator $s$ is also estimated with high accuracy. Only variant \textit{simple}
shows a larger error between the predicted and true contact points. The predicted
effective push velocity $\mathbf{v_p}$ also does not match the values we got from applying
the first stage of the analytical model on ground truth input very closely. Since these
errors do not seem to harm the overall prediction accuracy, we conclude that they cancel each 
other out. This shows that \textit{simple} is not
as strongly constrained by its analytical component as \textit{hybrid} and \textit{error} and
that it thus has more freedom in choosing its state representation.

\subsubsection*{Summary}

All our models reach the performance of the (perception-free) \textit{physics} baseline given 
enough training data. 
Combining neural networks and analytical models strongly improves performance in comparison to to 
purely learned models when little training data is available. However, 
\textit{neural} can achieve the highest prediction accuracy and beat the \textit{physics} baseline
when trained on a very large dataset.

To further improve the prediction accuracy of \textit{hybrid} while preserving its data-efficiency,
an additive error-correction term can be learned. Replacing a part of the analytical model 
with a learned component in \textit{simple} in contrast harmed the data efficiency.

\subsection{Generalization to New Pushing Angles and Contact Points}\label{sec:results-exp2}

The previous experiment showed the performance of the different models when 
testing on a dataset with a very similar distribution to the training set.
Here, we evaluate the performance of the networks on held-out push configurations 
that were not part of the training data. Note that 
while the test set contains \emph{combinations} of 
object pose and push action that the networks have not encountered during 
training, the pushing actions or object poses themselves do not lie outside of 
the value range of the training data. This experiment thus test the models' 
\emph{interpolation} abilities.

\subsubsection*{Data}

We again train the networks on a dataset that contains all objects and pushes 
with velocity $20\frac{mm}{s}$.
For constructing the test set, we collect all pushes with (i) pushing angles 
$\pm 20^{\circ}$ and $0^{\circ}$ to the
surface normal (independent from the contact points) and (ii) at a set of 
contact points illustrated in Figure~\ref{fig-data-a} 
(independent from the pushing angle).

The remaining pushes are split randomly into a training and a validation set, which 
we use to monitor the training process.
There are about 114k data points in the training split, 23k in the validation 
split and 91k in the test set.

\subsubsection*{Results}
As Table~\ref{tab-exp2} shows, \textit{hybrid} and \textit{error} perform best 
for predicting the object velocity for pushes that were not part of the training set. 
Although still being close, none of the networks can outperform the 
\textit{physics} baseline on this test set. 

Note that the difficulty of the 
test set in this experiment differs from the one in the previous experiment, as
can be seen from the different performance of the \textit{physics} baseline: 
Due to the central contact point locations and small pushing angles, the test set 
contains a high proportion of pushes with sticking contact (see Section~\ref{sec:model}),
for which the  resulting object movement is similar to the pusher movement. Prediction 
in sticking contact cases is therefore generally simpler than in cases in which the
pusher slides along the object.
This difference in difficulty makes it hard to compare the results between 
Table~\ref{tab-exp1} and Table~\ref{tab-exp2} in terms of absolute values.

With more than 100k training examples, we supply enough data for the pure 
neural model to clearly outperform the combined approach and the baseline in 
the previous experiment (i.e.\ when the test set is similar to the training 
set, see Figure~\ref{fig-datasize}). 
The fact that \textit{neural} now performs worse than \textit{hybrid} and 
\textit{physics} indicates that its advantage over 
the \textit{physics} baseline may not come from it learning a more accurate 
dynamics model. 
Instead, it probably memorizes specific input-output combinations that the 
analytical model cannot predict as well, e.g.\ due to noisy object pose 
data.

This might also be the reason why \textit{error} cannot improve on \textit{hybrid}
as much as in the previous experiment, especially when it comes to predicting the
translation of the object. It is however encouraging to see that the learned 
error correction term for the predicted rotation is still beneficial for pushes
not seen during training. 

In contrast to \textit{hybrid} and \textit{error}, \textit{simple} again does 
not seem to profit from using the simplified analytical model and 
performs similar to \textit{neural}. 

As in the previous experiment (see Table~\ref{tab-exp1}), we also tested 
the generalization ability of the networks when trained on a smaller training set. 
If we supply only 2500 training examples, the difference between \textit{hybrid} and
the purely learned model is again much more pronounced: \textit{Hybrid} achieves 
20.3\,\% translation and 43.8\,\% rotation error whereas \textit{neural} lies 
at 38.7\,\% and 63.4\,\% respectively. % for translation and rotation.

\begin{table}
\scriptsize
\caption{Prediction errors for testing on pushes with pushing angles and
contact points not seen during training. \label{tab-exp2}} 
\begin{tabular}{r  l l l} 
\toprule
 & trans  & rot   & pos [mm] \\
\midrule
\textit{neural} & $16.5 \: (0.06)$\,\% & $36.1 \: (0.17)$\,\% & $0.31 \: (0.001)$ \\
\textit{simple} & $16.4 \: (0.06)$\,\% & $37.1 \: (0.18)$\,\%  & $\mathbf{0.31 \: (0.001)}$ \\
\textit{hybrid}   & $\mathbf{15.6 \: (0.07)}$\,\% & $35.3 \: (0.19)$\,\% & $0.31 \: (0.001)$  \\
\textit{error}   & $\mathbf{15.6 \: (0.07)}$\,\% & $\mathbf{34.5 \: (0.18)}$\,\% & $0.32 \: (0.001)$  \\
\textit{neural dyn} & $18.1 \: (0.07)$\,\% & $44.1 \: (0.2)$\,\% & -\\
\midrule
\textit{physics}  &  $14.6 \: (0.06)$\,\% & $32.8 \: (0.18)$\,\% & -\\
%\midrule
\textit{zero} & $4.36 \: (0.013)$\, mm & $2.27 \: (0.009)\, ^{\circ}$ & - \\
\bottomrule
\end{tabular}
\end{table}

\subsubsection*{Summary}

The purely learned model performs worse than the hybrid approaches when interpolating 
to unseen push configurations. For all models,
the difference to the \textit{physics} baseline is larger when the training distribution
does not match the test distribution.

\subsection{Generalization to Different Push Velocities}\label{sec:results-exp3}

In this experiment, we test how well the networks generalize to unseen push 
velocities. In contrast to the previous experiment, the test actions 
in this experiment have a different value range than the actions in the 
training data, and we are thus looking at \emph{extrapolation}. 
As neural networks are usually not good at extrapolating beyond 
their training domain, we expect the model-based network variants 
to generalize better to push-velocities not seen during training.

\subsubsection*{Data}

We use the networks that were trained in the first experiment 
(\ref{sec:results-exp1}) on the full (190k) training set. The push velocity in the 
training set is thus 20\,$\frac{mm}{s}$. We evaluate on datasets with different push 
velocities ranging from 10\,$\frac{mm}{s}$ to 300\,$\frac{mm}{s}$.
Since seeing only one push velocity during training might be a disadvantage
for the learned models, we also compose two new training datasets, one with 
velocities conform to the quasi-static assumption (10 and 20\,$\frac{mm}{s}$)
and one with a higher second velocity (20 and 50\,$\frac{mm}{s}$) that 
violates the quasi-static assumption. Both datasets have slightly more than 125k
training examples.

\subsubsection*{Results}

\begin{figure*}
\centering
\includegraphics[width=\linewidth]{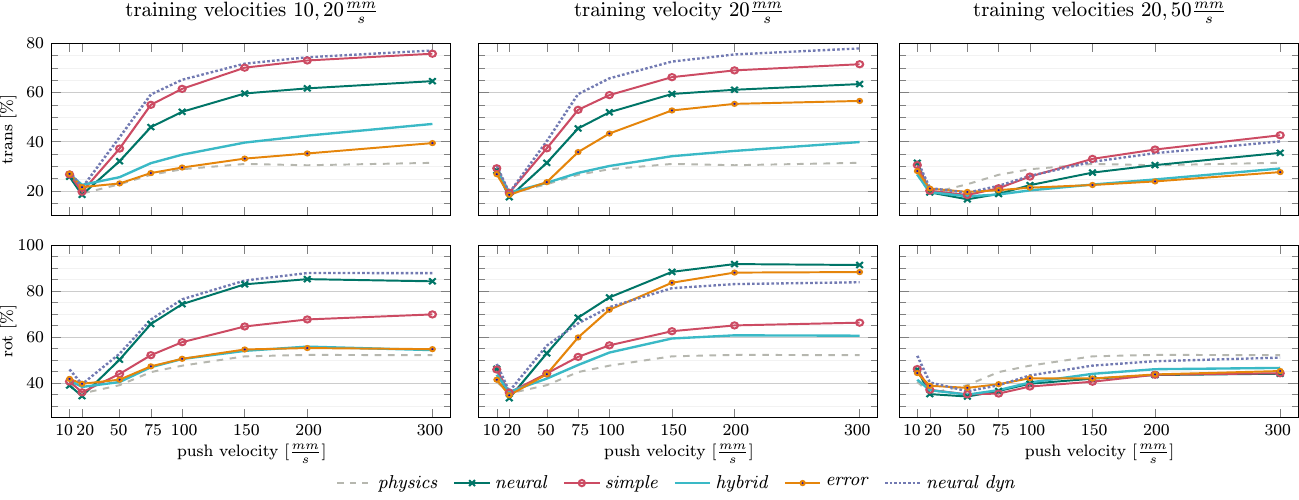}
\caption{Errors on predicted translation and rotation for testing on different 
push velocities. In the first column, all models were trained on push velocities 10 and 
20\,$\frac{mm}{s}$, in the second column on velocity 20\,$\frac{mm}{s}$ and in the
last column on velocities 20 and 50\,$\frac{mm}{s}$. 
When training on velocities that are small enough to ensure quasi-static pushing,
all models have trouble extrapolating to higher velocities, but 
\textit{hybrid} and \textit{error} stay much closer to the \textit{physics} baseline
than \textit{simple}, \textit{neural} and \textit{neural dyn}. 
Seeing additional training data from a higher push velocity (50\,$\frac{mm}{s}$) that 
violates the quasi-static assumption strongly improves the generalization to higher 
velocities for all models and enables them to beat the \textit{physics} baseline in many cases.
Especially for the predicted object translation, we however still see a much stronger decrease
in performance for \textit{simple}, \textit{neural} and \textit{neural dyn} than for
\textit{hybrid}, \textit{error} and \textit{physics}.
\label{fig-vel}}
\end{figure*}

%\begin{figure}
%\centering
%\includegraphics[width=\linewidth]{plot_vel}
%\caption{Errors on predicted translation and rotation for testing on different 
%push velocities. All models were trained on push velocity 20\,$\frac{mm}
%{s}$.  While \textit{hybrid} stays close to the \textit{physics} baseline, the other 
%models have trouble extrapolating. 
%\label{fig-vel}}
%\end{figure}

Results are shown in Figure~\ref{fig-vel}. Since the input action does not 
influence perception of the object position, we only 
report the errors on the predicted object motion. 

When training on push velocities below 50\,$\frac{mm}{s}$,
we see a very large difference between the performance of 
our combined approach and the pure neural network for higher velocities. 
\textit{Neural's} and \textit{neural dyn's} predictions quickly become very 
inaccurate, with the error on predicted translation rising to more than 
60\,\% and the error on predicted rotation to more than 80\,\% of the error when 
predicting zero movement always. The performance of \textit{hybrid} on the other hand 
is most constant over the different push velocities and declines only slightly more than 
the \textit{physics} baseline. \textit{Error}, too, extrapolates well, but only when
trained on more than one push velocity. 

Like \textit{neural} and \textit{neural dyn}, \textit{simple}, too, 
gets worse on higher velocities. Its performance when predicting rotations however 
degrades much less than for predicting translations. The reason for this 
is that all three architectures 
struggle mostly with predicting the correct \emph{magnitude} of the object 
translation and not so much with predicting the translation's 
\emph{direction}. By using the second stage of the analytical model, 
\textit{simple} has information about how the direction of the object 
translation and the contact point relate to its rotation, which results in 
much more accurate predictions. 

The advantage of \textit{hybrid} for extrapolation lies in the 
first stage of the analytical model, which allows it to scale its predictions
according to the magnitude of the action and the contact indicator $s$. 
Both are in essence multiplication operations. A general multiplication of 
inputs can however not be expressed using only fully-connected layers (as used by \textit{simple}, 
\textit{neural}, \textit{neural dyn} and the error-prediction part of \textit{error})
because fully-connected layers essentially perform weighted additions of 
their inputs. So instead of learning the underlying function, the networks 
are forced to resort to memorizing input-output relations for the magnitude
of the object motion, which explains why extrapolation does not work well,
especially when training on low push velocities. 

When combining the prediction of the analytical model with a learned error-term
and training only on one push velocity, the resulting model suffers from the 
same issues as the other network-based variants. The decline is however less pronounced 
than for \textit{neural}, and only starts after $50\frac{mm}{s}$.
A possible reason for this is that the error-correction term is rather 
small compared to the output of the analytical model. This means that the weights 
with which the action enters the computation of the error term are smaller
than for \textit{neural}, \textit{simple} or \textit{neural dyn}. 
 
Interestingly, adding a second training velocity completely changes the picture
and makes \textit{error} perform on par with or even better than \textit{hybrid}. Our hypothesis
is that seeing different velocities during training prevents the error term from
overfitting to the input action and minimizes the effect of the action magnitude on 
the predicted error-term. 
In Section~\ref{sec:results-exp5}, we show that for training on only one push velocity,
\textit{error} can also be made more
robust to higher velocities by normalizing the push action before using it 
as input to the error-prediction.
%The hypothesis that \textit{error} overfits when trained
%on only one velocity is also substantiated by experiments with shorter training times:
%After 60k training steps, \textit{error} trained on $20 \frac{mm}{s}$ pushes extrapolates much better 
%to higher velocities than after the full 75k, despite of showing a higher validation loss.

While the \textit{physics} baseline performs better than the models trained on low push velocities, 
it predictions also get worse on higher push velocities.
The main reason for this is that the quasi-static assumption of the model is violated: For 
pushes faster than 20\,$\frac{mm}{s}$, the object gets accelerated and can 
continue sliding even after contact to the pusher was lost. 

How different the dynamics of pushing are between the quasi-static and this dynamic behaviour
also becomes apparent when we include the push velocity 
50\,$\frac{mm}{s}$ in the training data for our learned models: They all extrapolate much better 
to higher velocities and are often able to outperform the \textit{physics} baseline. This increase 
of performance for fast pushes however only extends to the slowest push velocity 
(10\,$\frac{mm}{s}$) for \textit{hybrid}, whereas all other models perform slightly 
worse than their counterparts that were only trained on one push velocity.

We also still see that with increasing push velocities, the variants
\textit{simple}, \textit{neural} and \textit{neural dyn} make significantly larger 
errors for predicting the translation of the object than \textit{hybrid} and \textit{error}.
Interestingly, for predicting the object rotation, all models except for \textit{neural dyn}
perform extremely well, with \textit{hybrid} even doing slightly worse than the others. 
A possible
reason for this difference between translation and rotation could be that the magnitude of 
rotations does not increase as strongly with the push velocity as the magnitude of translations:
The average rotation increases from 1.4$^{\circ}$ on 10\,$\frac{mm}{s}$ pushes to 
14.4$^{\circ}$ on 300\,$\frac{mm}{s}$ pushes, whereas translation increases from 2.1 to 24.8\,mm.
The models therefore need to change their predicted rotations less in response
to higher push velocities than they have to for translation.

\subsubsection*{Summary}

Extrapolating to different push velocities is difficult for purely learned models,
especially when the training data only contains low pushing velocities. Using the 
analytical model in \textit{hybrid} and \textit{error} facilitates extrapolation by 
providing multiplication operations and explaining the influence of the action
on the resulting movement.
Since the quasi-static assumption of the analytical model is violated by fast pushes,
our models can however learn to outperform the \textit{physics} baseline in this regime  
when they have training data from faster pushes.

\subsection{Generalization to Different Objects}\label{sec:results-exp4}

This experiment tests how well the networks generalize to unseen
object shapes and how many different objects the networks have 
to see during training to generalize well.

\subsubsection*{Data}

We train the networks on three different datasets: With one object (butter), 
two objects (butter and hex) and three objects (butter, hex and
one of the ellipses or triangles). The datasets with
fewer objects contain more augmented data, such that the total number 
of data points is about 35k training examples in each. As test sets, we use one 
dataset containing the three ellipses and one containing all triangles.
While this is fewer training data than in the 
previous experiments, it should be sufficient for the pure neural network 
to perform as well as \textit{hybrid}, since the test sets contain only few 
objects.
 
\subsubsection*{Results}
\begin{figure*}[t]
\centering
\includegraphics[width=\linewidth]{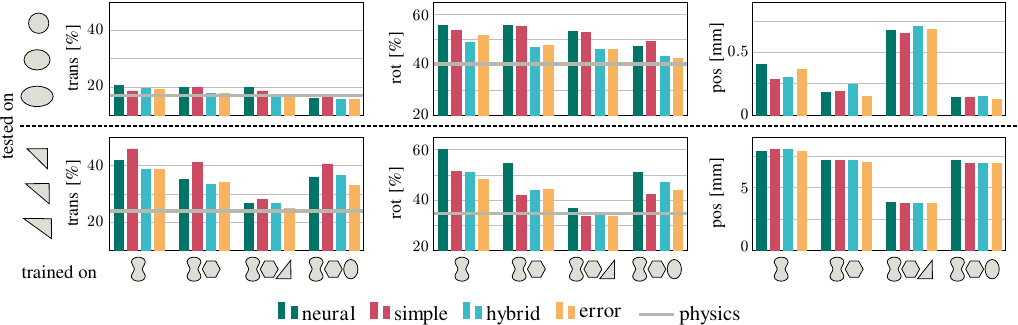}
\caption{Prediction errors in translation, rotation and
position on objects not seen during training. Training objects are shown on the $x$-axis. 
The top row shows results for evaluating on ellipses, the bottom row on 
triangles. All networks generalize well to ellipses, but are worse for 
triangles, where the error in predicted position is by factor ten 
higher than for the other objects. \textit{Neural} particularly struggles with predicting
rotations of previously unseen objects.
\label{fig-general} 
}
\end{figure*}

The results in Figure~\ref{fig-general} show that \textit{neural} is 
consistently worse than the other networks when predicting rotations. It also 
improves most notably when one example of the test objects is in the training 
set. The differences between the models are less pronounce when predicting translation, 
except for \textit{simple} which performs particularly bad on triangles. 
The different models do not differ much when predicting position, which is not 
surprising, since they share the same perception architecture. The architecture
with added error-term does not perform very different from \textit{hybrid}, which 
implies that the error-correction term does not depend much on the shape of the
object.

In general, all models perform surprisingly well on ellipses, even if the 
models only had access to data from the butter object. Reaching the baseline
performance on triangles is however only possible with a triangle in the 
training set. 
Predicting the object's position is most sensitive to the shapes seen during 
training: It generalizes well to ellipses which have similar shape and 
size as the butter or hex object. The triangles on the other hand are very different 
from the other objects in the dataset and the error for localizing triangles is by 
factor ten higher than for ellipses. 

\subsubsection*{Summary}

Using the analytical model in \textit{hybrid} and \textit{error} also facilitates 
generalization to novel object shapes, which is more difficult for the purely learned model.
All models struggle slightly with localizing objects of unknown shapes.

\subsection{Visualizations}

As a qualitative evaluation, we plot the predictions of our networks, the \textit{physics} 
and \textit{neural dyn} baselines and the ground truth object motion for 200 repetitions of 
the same push configuration. The data for these repeated pushes is available with the MIT Push dataset. 
All repetitions have the same nominal pushing angle ($0^{\circ}$), velocity 
($20 \frac{mm}{s}$) and contact point, but the exact values vary slightly between individual pushes. 
To keep the visual input diverse, we also sample a different transformation of 
the whole scene for each repetition, such that the object's pose in the image varies. 
The networks were trained on the full dataset from Experiment~\ref{sec:results-exp1}.

\begin{figure*}
\centering
\begin{subfigure}{0.245\textwidth}
\includegraphics[width=\textwidth]{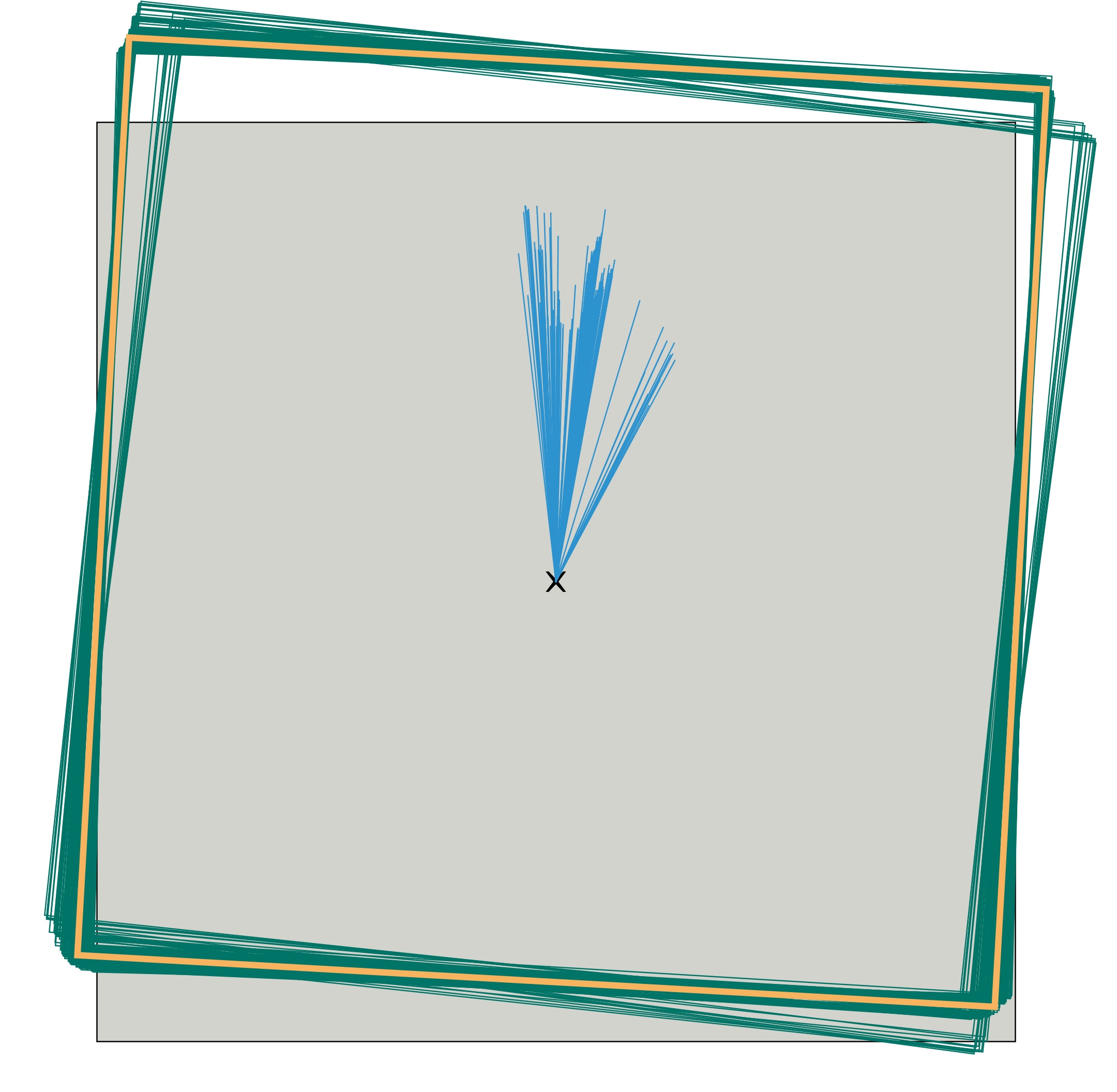}
\caption{Ground truth}
\end{subfigure}
\begin{subfigure}{0.245\textwidth}
\centering
\includegraphics[width=\textwidth]{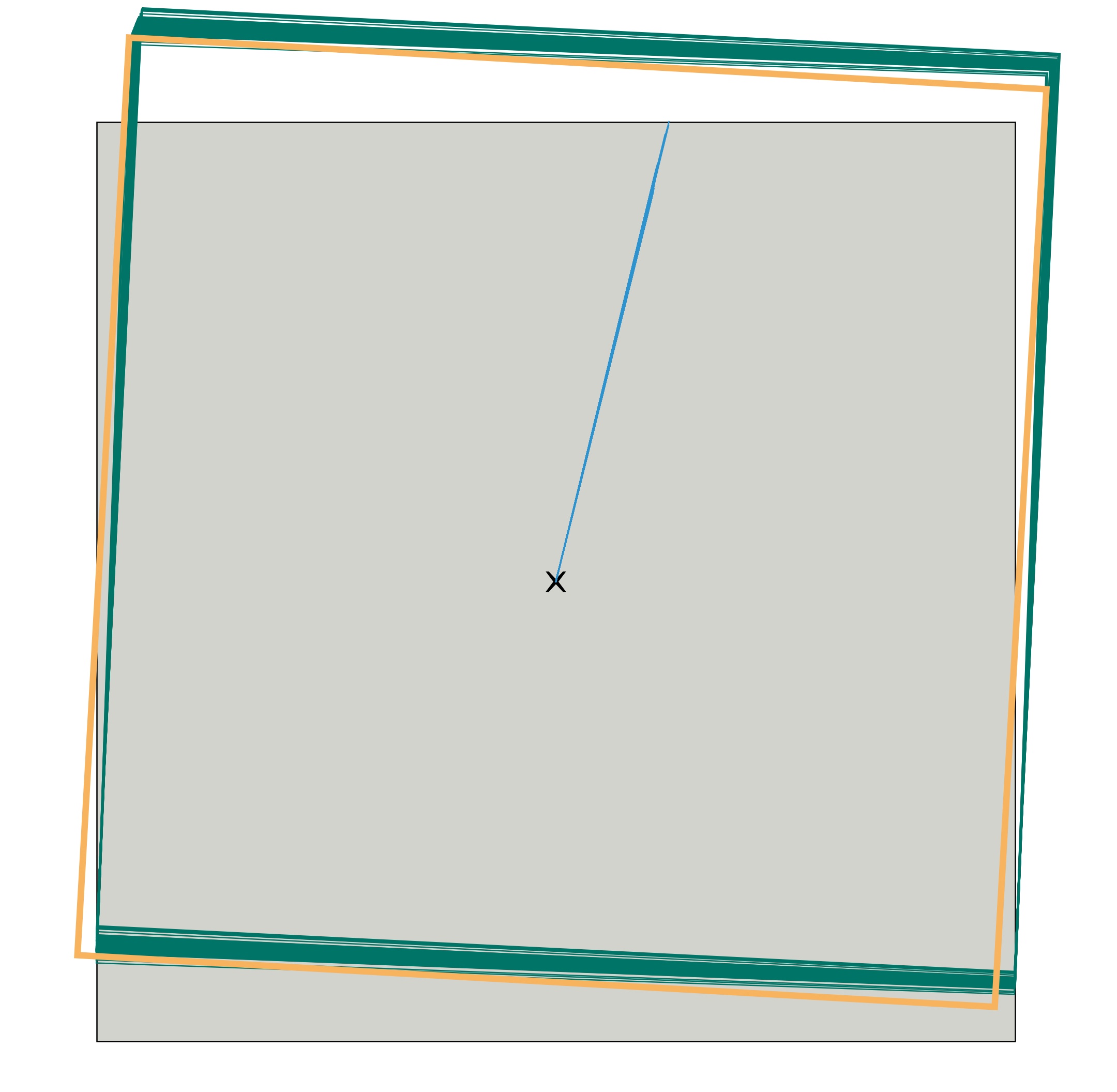}
\caption{\textit{Physics} \citep{model}}
\end{subfigure}
\begin{subfigure}{0.245\textwidth}
\centering
\includegraphics[width=\textwidth]{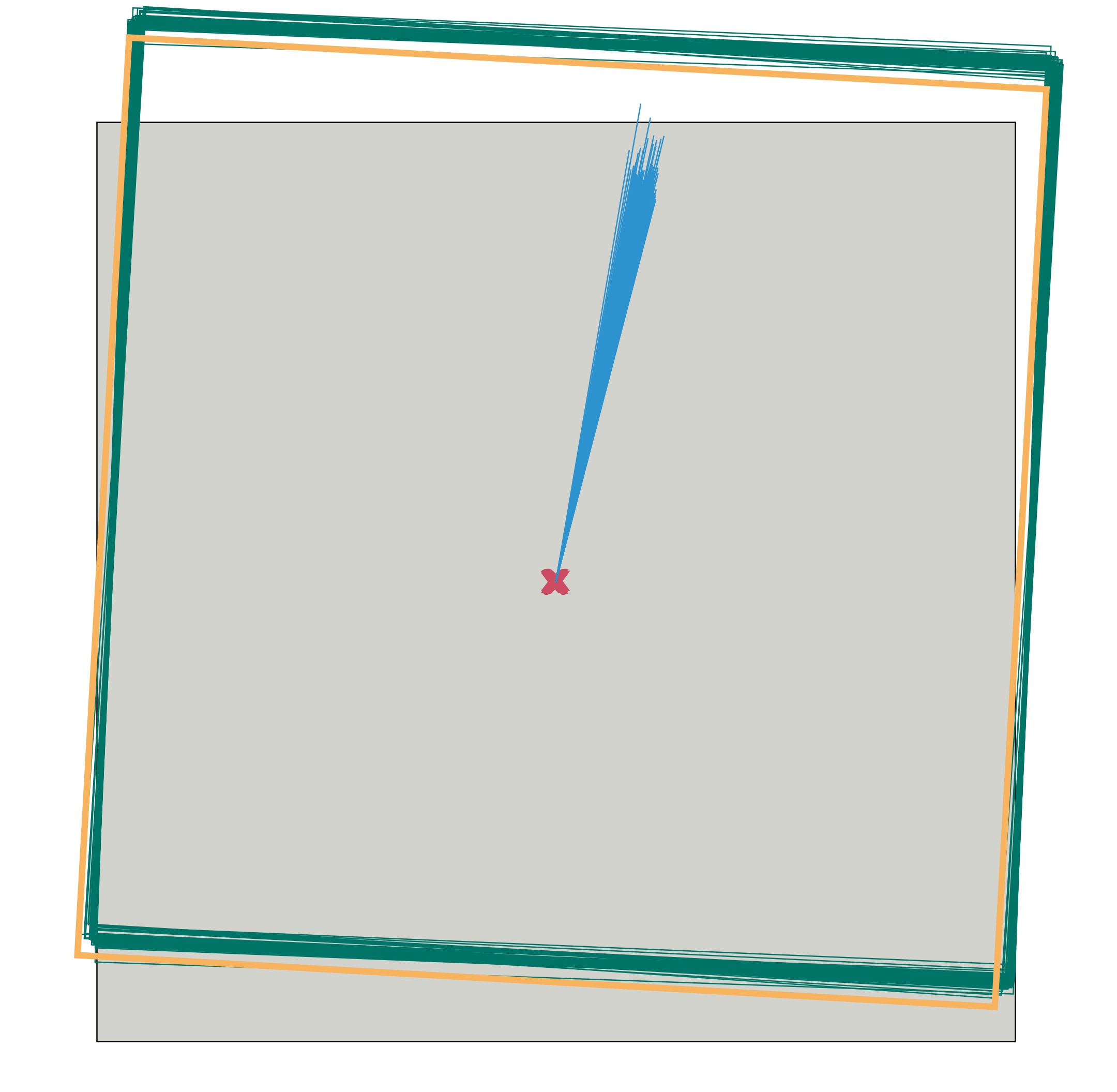}
\caption{\textit{Hybrid}}
\end{subfigure}
\begin{subfigure}{0.245\textwidth}
\includegraphics[width=\textwidth]{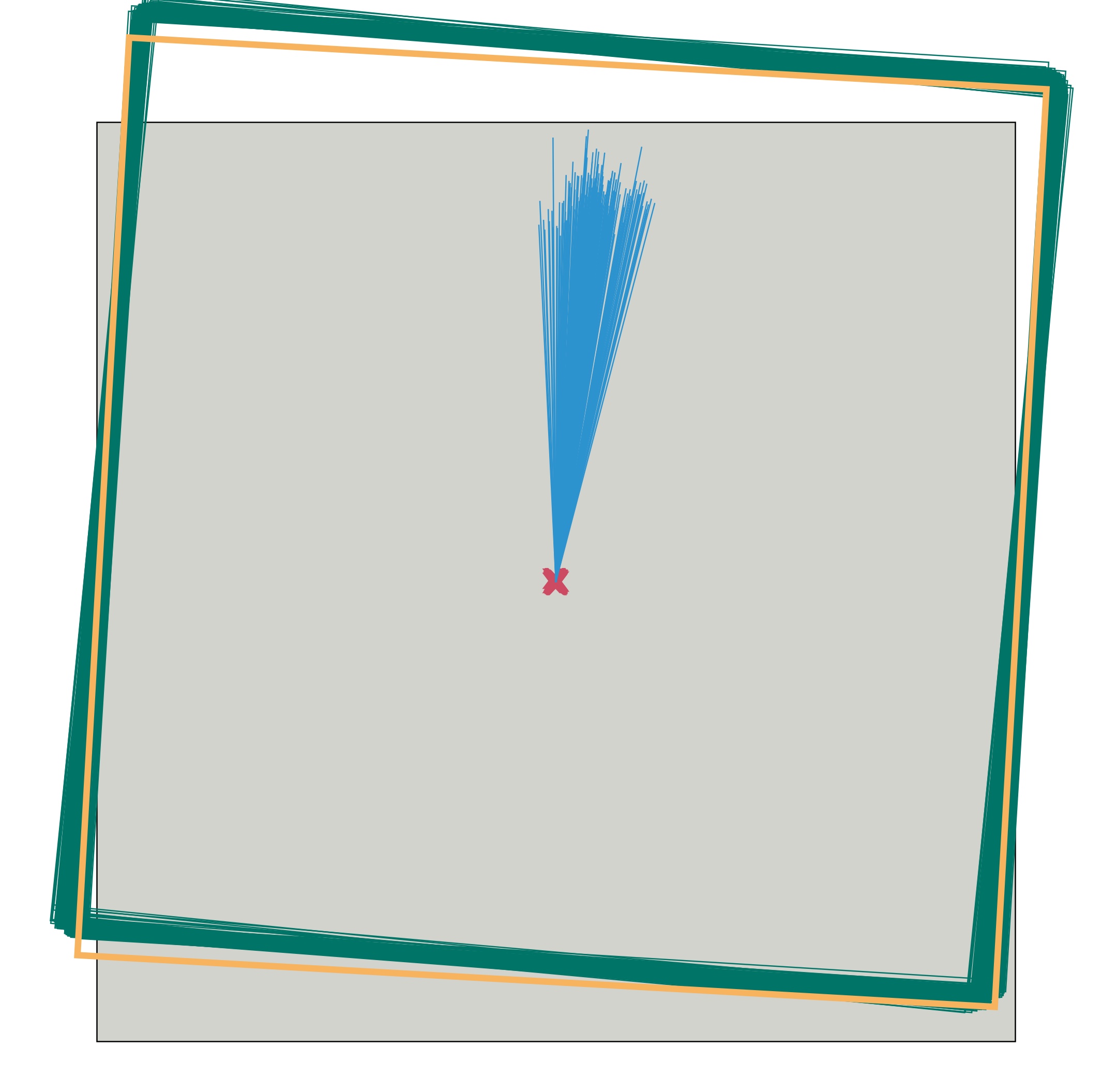}
\caption{\textit{Simple}}
\end{subfigure}
\\
\begin{subfigure}{0.245\textwidth}
\includegraphics[width=\textwidth]{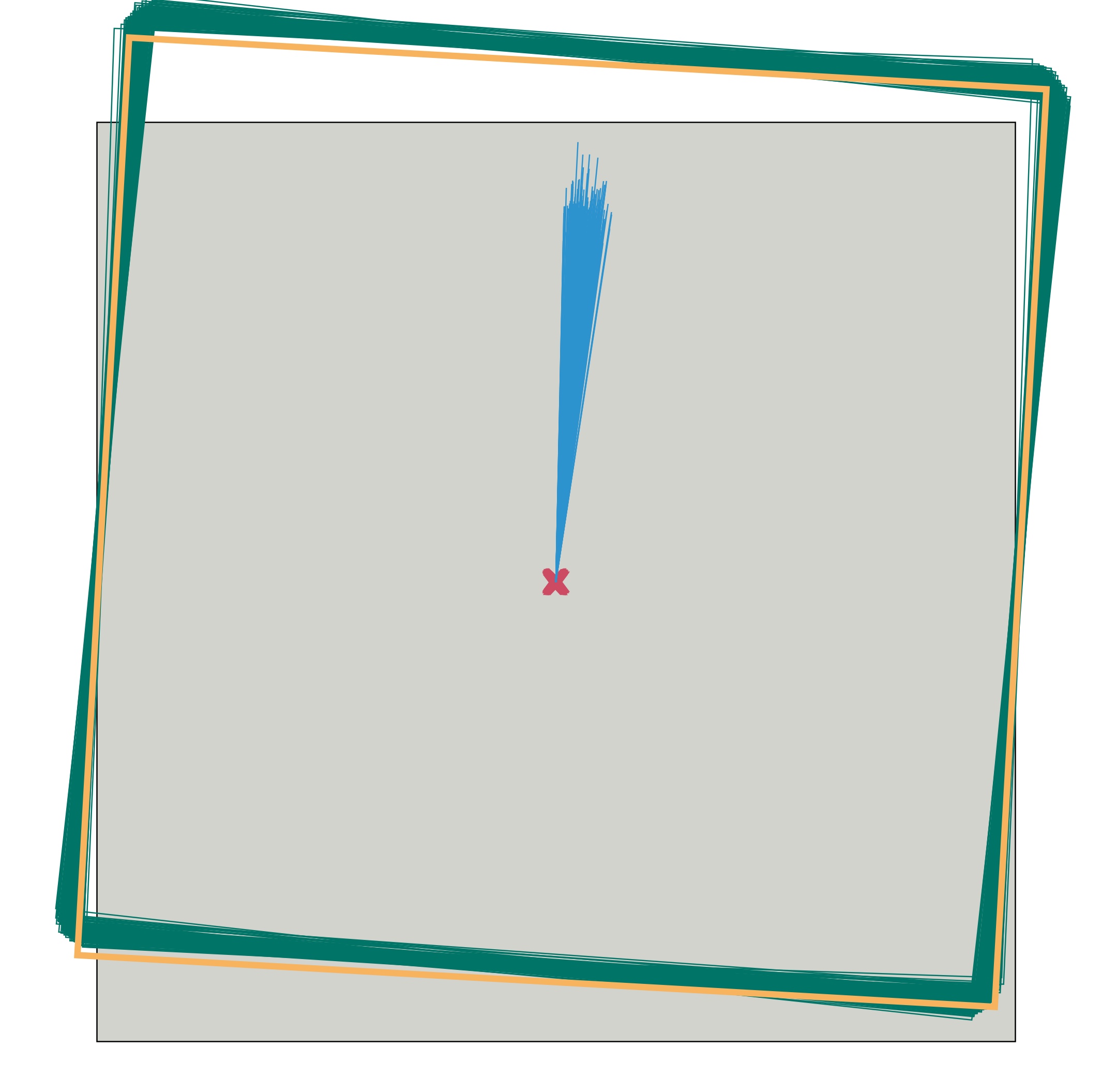}
\caption{\textit{Neural}}
\end{subfigure}
\begin{subfigure}{0.245\textwidth}
\includegraphics[width=\textwidth]{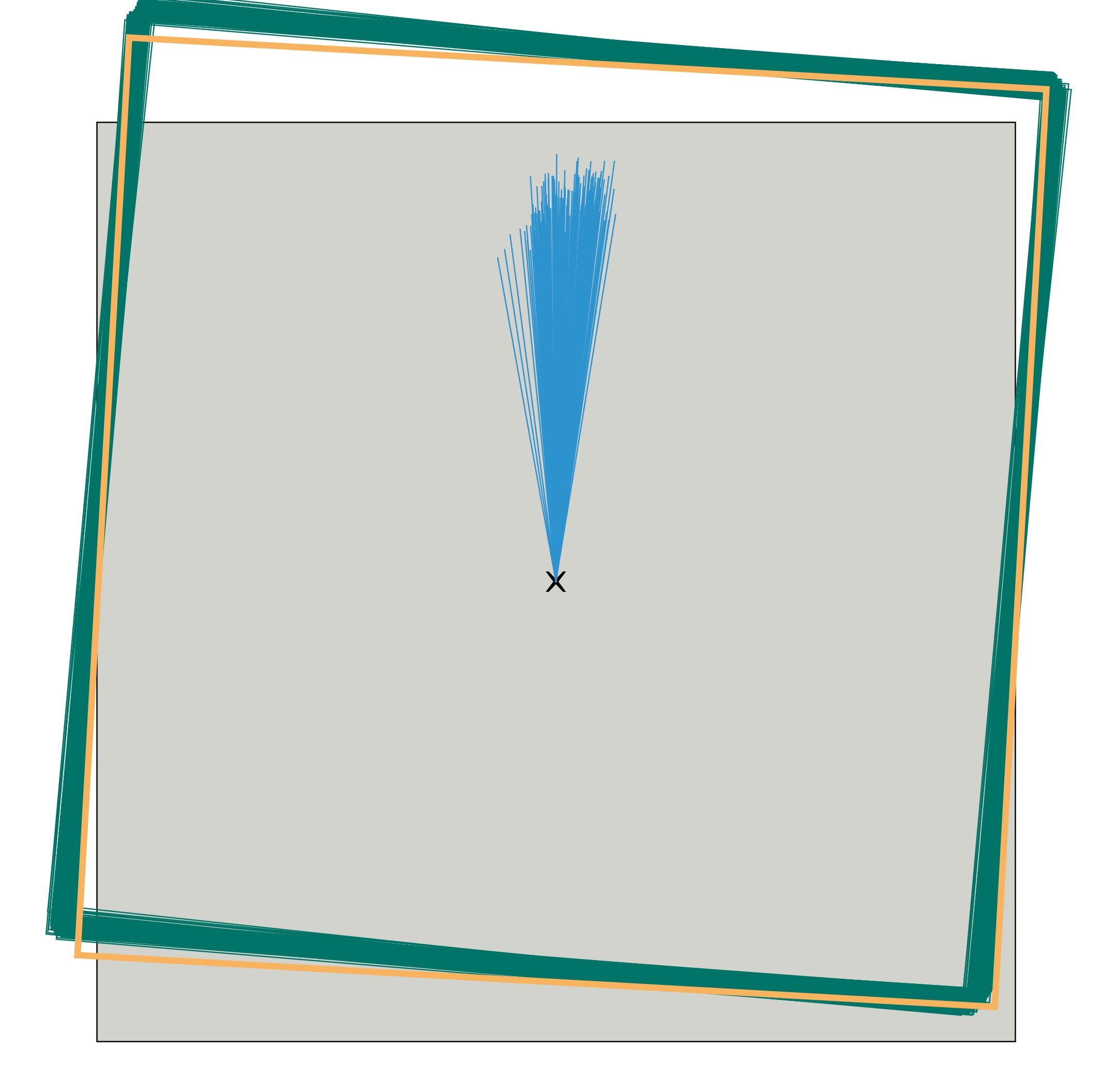}
\caption{\textit{Neural dyn}}
\end{subfigure}
\begin{subfigure}{0.245\textwidth}
\includegraphics[width=\textwidth]{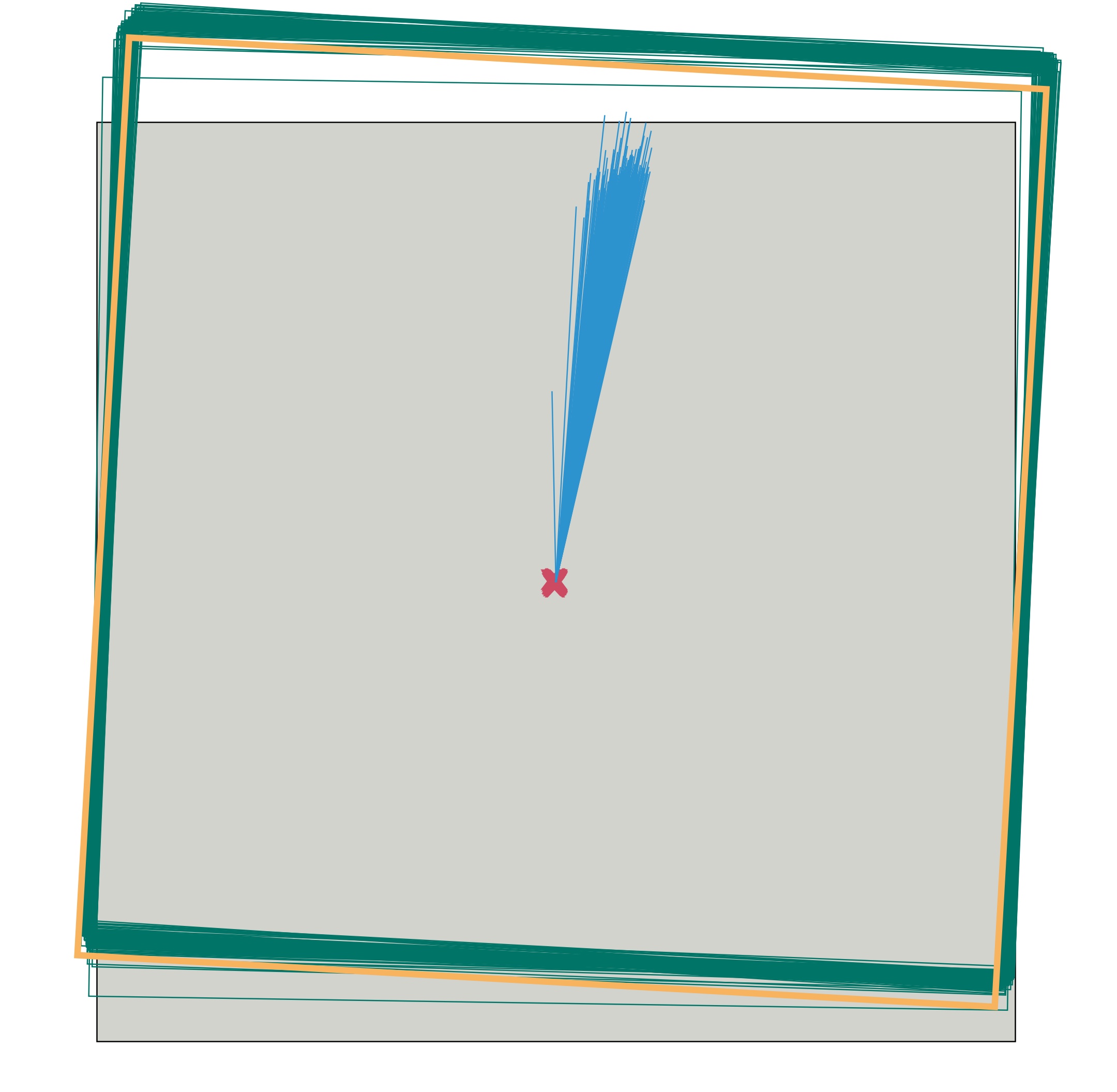}
\caption{\textit{Error}}
\end{subfigure}
\caption{Qualitative evaluation on 200 repeated pushes with the same push configuration 
(angle, velocity, contact point). The green rectangles show the (predicted) 
pose of the object after the push and the blue lines illustrate the object's translation 
(for better visibility, we upscaled the lines by factor 5). The thicker orange rectangle 
is the average ground truth pose of the object after the push. Red crosses indicate the 
predicted initial object positions. All models
predict the movement of the object and its initial position well, but cannot capture the
multimodal distribution of the ground truth data.}
\label{fig-variance}
\end{figure*}

The results shown in Figure~\ref{fig-variance} illustrate that the resulting ground truth
object motion for the same push configuration varies greatly between trials. Especially in 
terms of object rotation, the distribution of outcomes shows two distinct modes (one close to the
overall mean and one with notably stronger object rotation). By comparing the ground truth
with the prediction of the analytical model, we can estimate how much of this variance is due 
to slight changes in the push configuration between trials (these also reflect in the analytical 
model) and how much is caused by other, non-deterministic effects.

\begin{figure*}
\centering
\begin{subfigure}{0.19\textwidth}
\includegraphics[width=\textwidth]{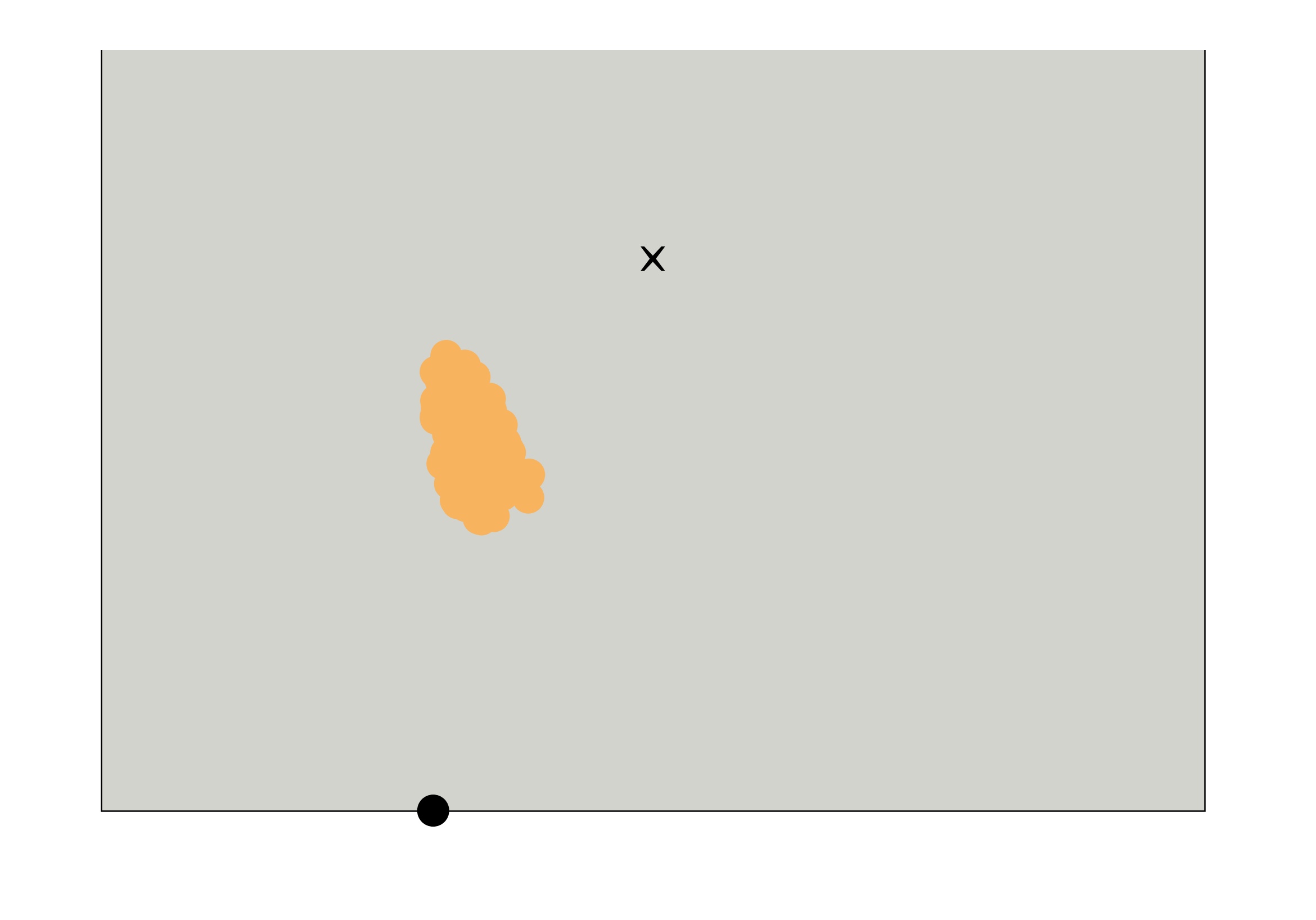}
\caption{Contact points predicted by \textit{simple}}
\end{subfigure}
\begin{subfigure}{0.19\textwidth}
\includegraphics[width=\textwidth]{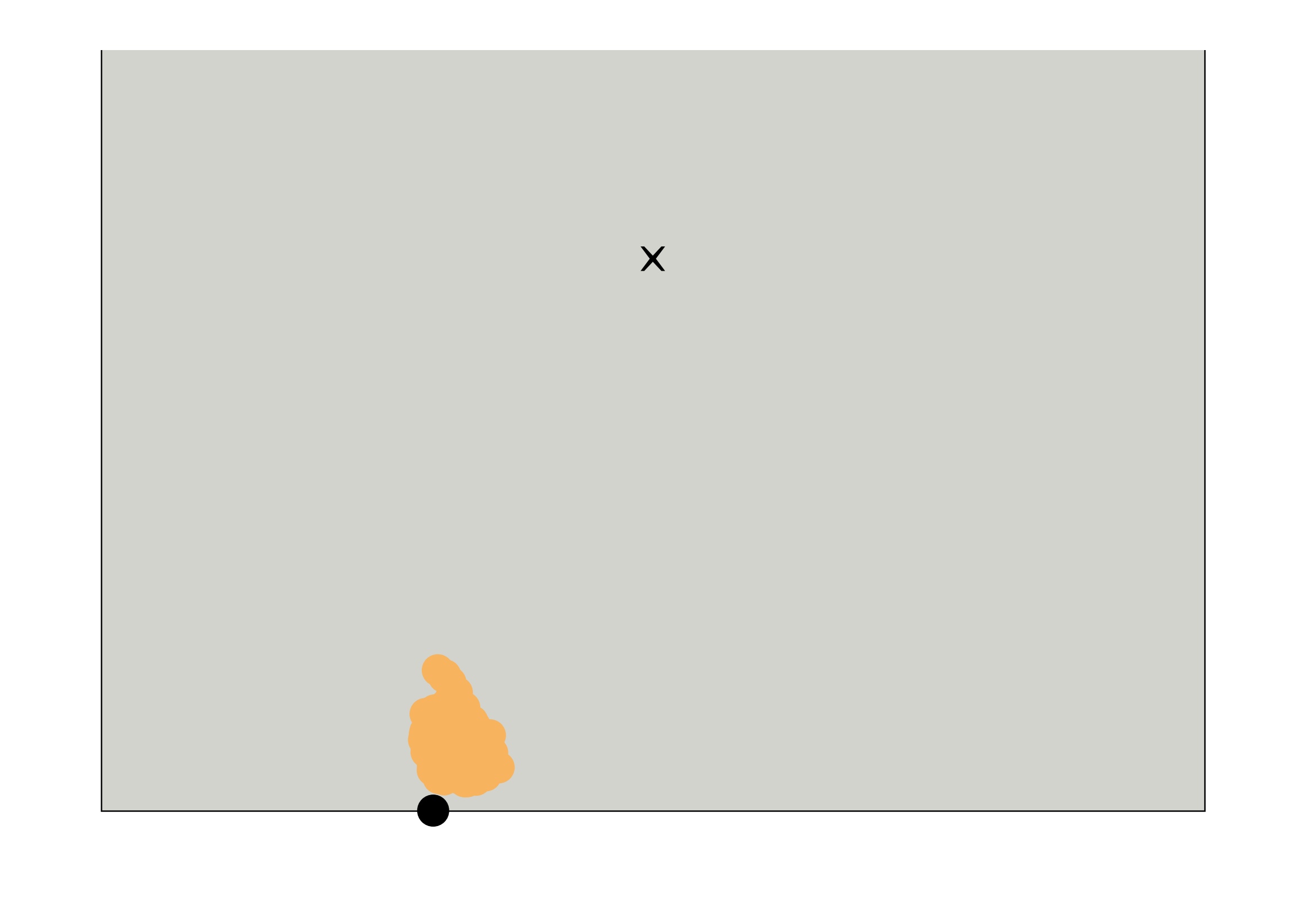}
\caption{Contact points predicted by \textit{hybrid}}
\end{subfigure}
\begin{subfigure}{0.19\textwidth}
\includegraphics[width=\textwidth]{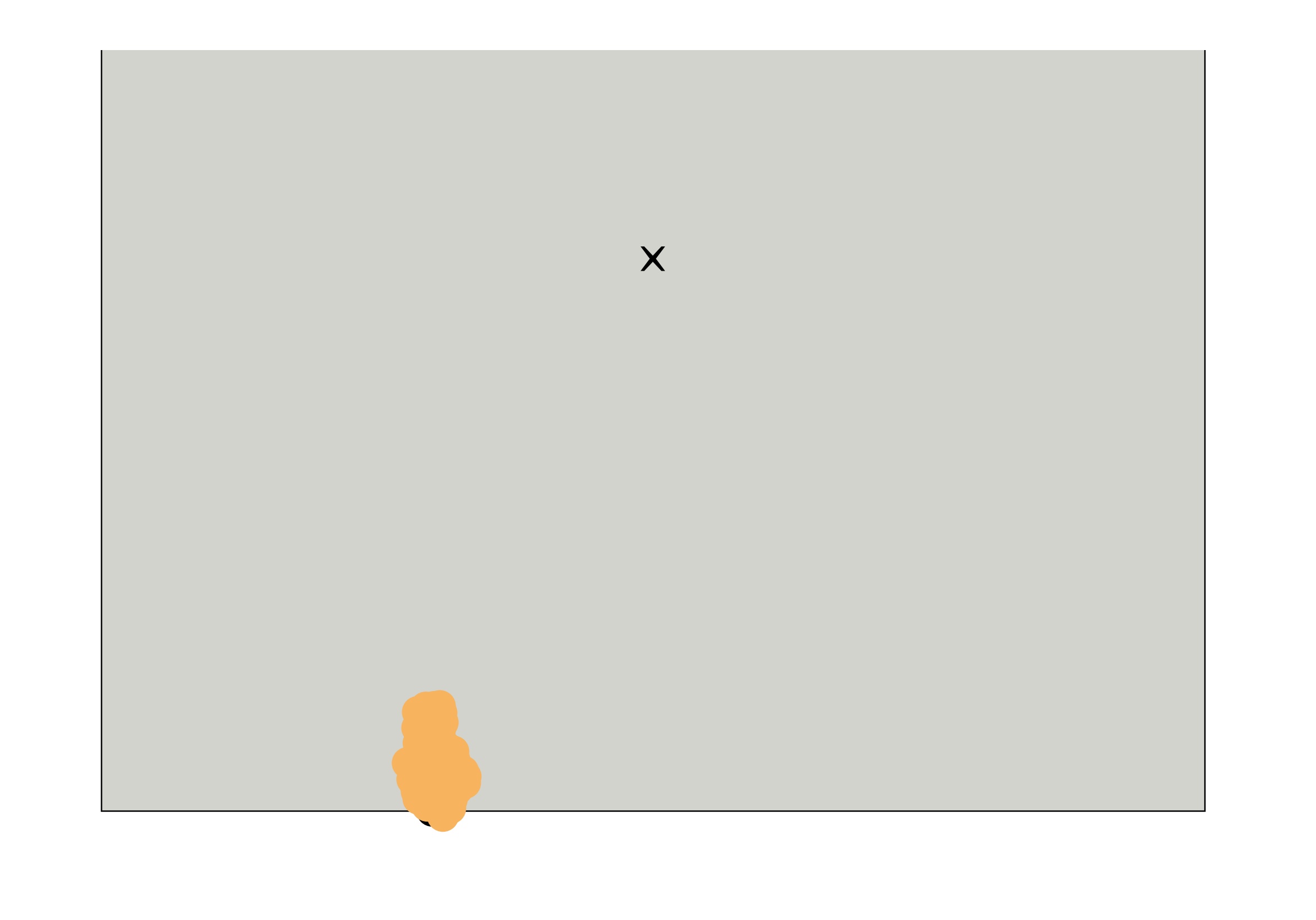}
\caption{Contact points predicted by \textit{error}}
\end{subfigure}
\begin{subfigure}{0.19\textwidth}
\includegraphics[width=\textwidth]{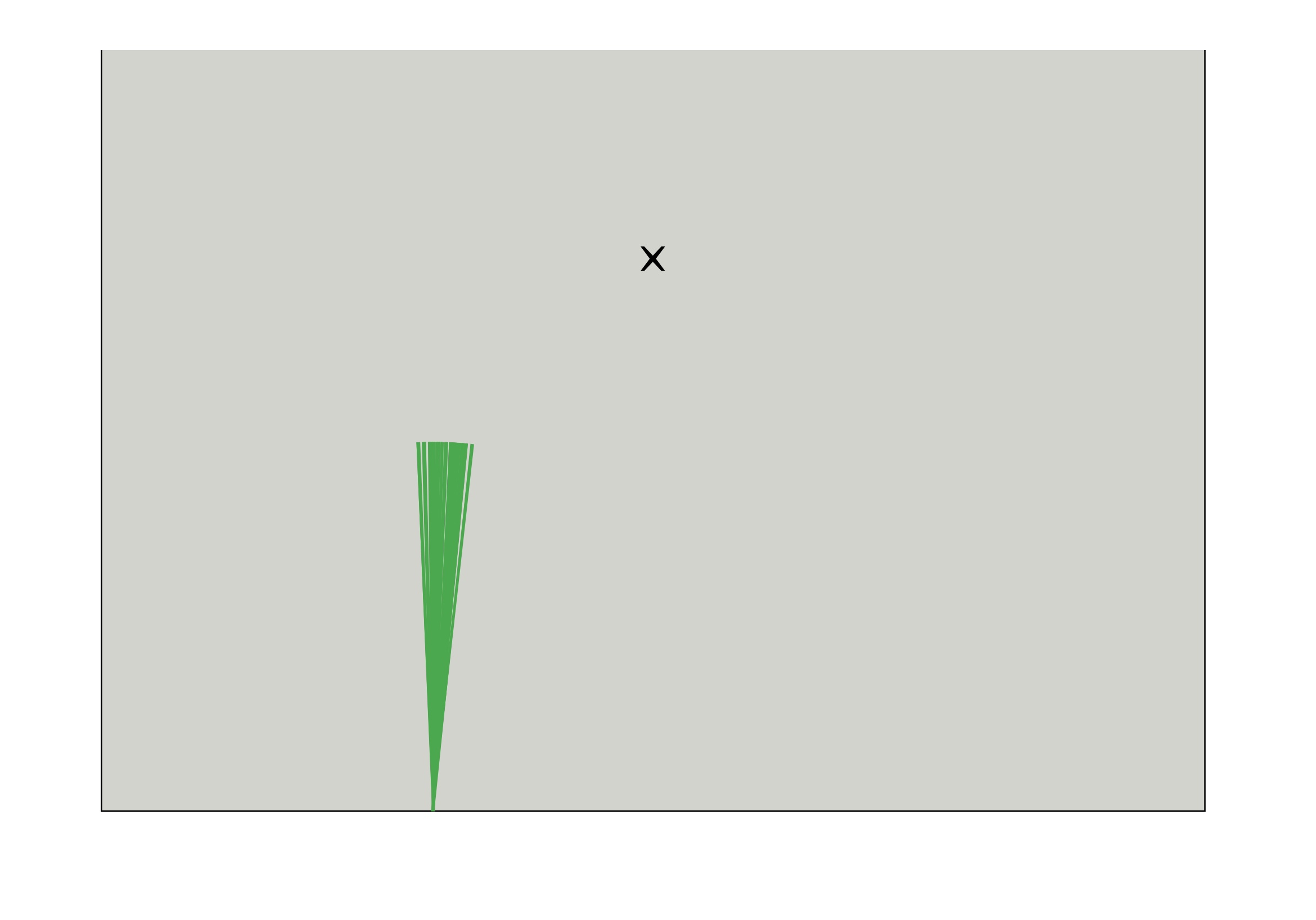}
\caption{Contact normal predicted by \textit{hybrid}}
\end{subfigure}
\begin{subfigure}{0.19\textwidth}
\includegraphics[width=\textwidth]{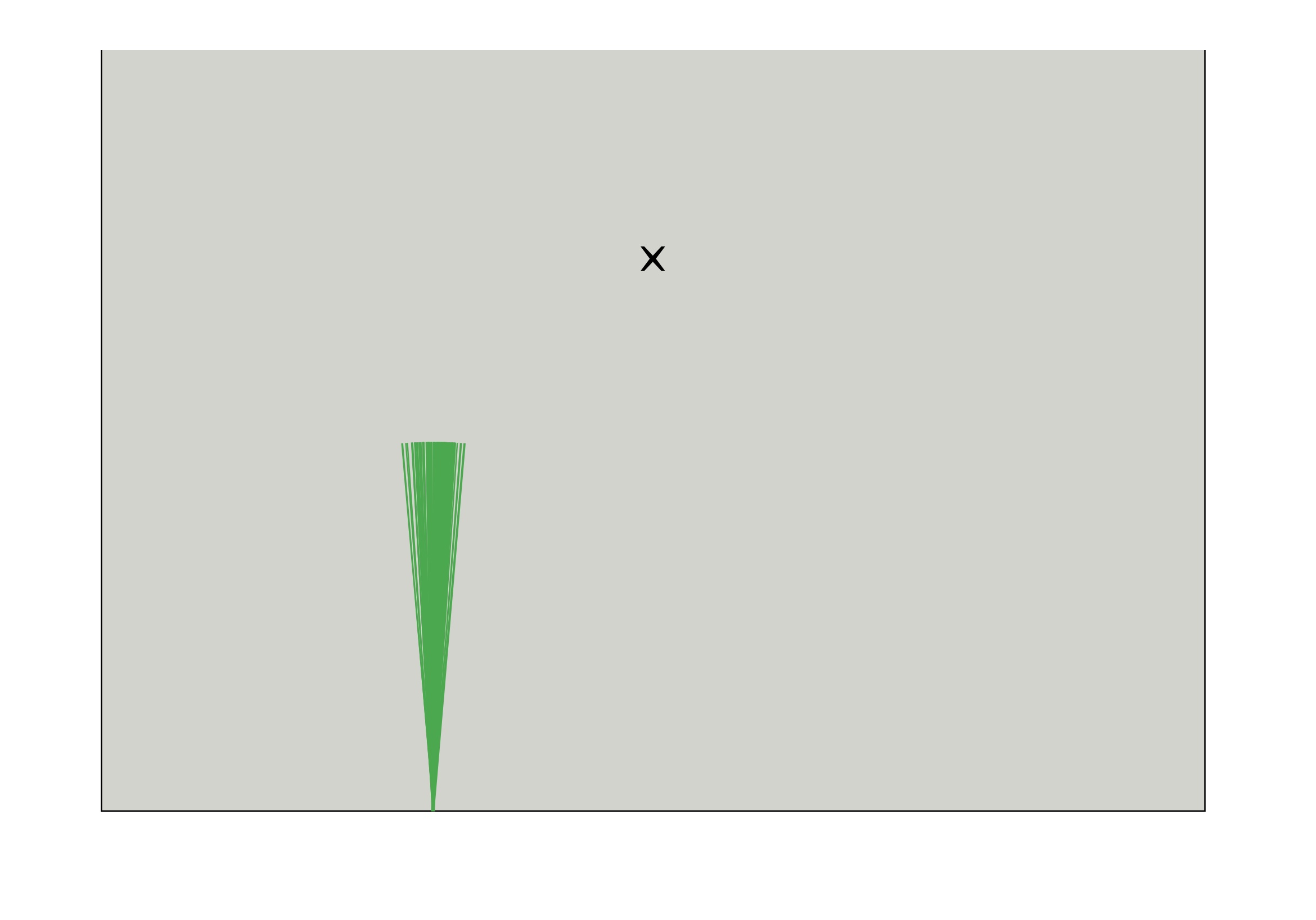}
\caption{Contact normal predicted by \textit{error}}
\end{subfigure}
\caption{Predicted contact points and normals from 200 repeated pushes with the same push 
configuration (angle, velocity, contact point). The black point marks the (average) ground 
truth contact point. While \textit{hybrid} and \textit{error} make fairly accurate 
predictions, \textit{simple} predicts the contact points not on the edge of the object but close to its center.}
\label{fig-contact}
\end{figure*}

The predictions of \textit{hybrid} and the analytical model are very similar. This again shows
that the state-representation that the neural network part of \textit{hybrid} predicts is
mostly accurate. The plotted contact point and normal estimates in Figure~\ref{fig-contact} 
further confirm this. Adding an error-correction term to the \textit{hybrid} architecture improves 
the average estimation quality a little, but also increases the variance of the predictions.

The visualizations for the other models (Figure~\ref{fig-variance} (d)-(f)) show that they, too, make 
good predictions in this example, but \textit{simple}
and \textit{neural dyn} have much more variance in the direction of the predicted translation than 
\textit{hybrid} or \textit{neural}. It is also interesting to see that \textit{neural}, 
\textit{neural dyn} and \textit{simple} all slightly overestimate the object rotation in
comparison to the mean ground truth movement, whereas \textit{physics} slightly underestimates it.
Figure~\ref{fig-contact} also shows that \textit{simple} is not very accurate in predicting 
the contact points, confirming the quantitative results found in Table\ref{tab-exp1-state}. As 
stated before, we believe that this inaccuracy is compensated for by the predicted $\mathbf{v_p}$.

\subsection{Evaluation of the Error-Architecture}\label{sec:results-exp5}

The previous results have shown that adding a learned error-correction term
to the output of the analytical model in the hybrid architecture enables the network
to improve over the performance of the analytical model.  
The \textit{error} model we analysed is able to outperform \textit{hybrid} and the 
\textit{physics} baseline if the training set and the test set are similar 
(see Experiment~\ref{sec:results-exp1}). 

In the following experiments, we evaluate different choices we made for the architecture
of \textit{error}. We also compare the ability of \textit{hybrid} and \textit{error} to 
compensate for larger errors in the analytical model.

\subsubsection*{Evaluation of Different Architectures}

As explained in Section~\ref{sec:networks}, we chose to block the propagation of gradients 
from the error-correction module to the glimpse-encoding, because we did not want the 
error-computation to interfere with the prediction of the state representation. Here, we 
also evaluate an architecture \textit{err-grad} that does not block the gradient propagation.
This architecture manages to beat \textit{hybrid} by an even bigger margin, as shown in Table 
\ref{tab-error-exp1}. 

\begin{table}
\scriptsize
\caption{Evaluation of different architectures for predicting an error-correction term.
In contrast to \textit{error}, \textit{error-grad} allows the propagation of gradients from 
the error-prediction module to the glimpse encoding. \textit{Error-norm} instead normalizes 
the push action to unit length before using it as input to the error-prediction.
Values shown are for training on the full training set (190k examples). Results for \textit{hybrid} and \textit{neural} are repeated for reference. 
\label{tab-error-exp1}} 
\begin{tabular}{r l l l} 
\toprule
  & trans  & rot   & pos [mm] \\
\midrule
\textit{neural} & $\mathbf{17.4 \: (0.12)}$\,\% & $\mathbf{33.4 \: (0.28)}$\,\% & $0.31 \: (0.002)$ \\
\textit{hybrid}  & $19.3 \: (0.13)$\,\% & $36.1 \: (0.3)$\,\% & $0.32 \: (0.002)$ \\
\textit{error}  & $ 18.4 \: (0.12)$\,\% & $ 34.6\: (0.29)$\,\% & $ 0.31\: (0.002)$  \\
\textit{error-grad}  & $ 17.9 \: (0.12)$\,\% & $ 34.4\: (0.29)$\,\% & $ \mathbf{0.29\: (0.002)}$  \\
\textit{error-norm}  & $ 18.3 \: (0.12)$\,\% & $ 35.3\: (0.29)$\,\% & $ 0.31\: (0.002)$  \\
\midrule
\textit{physics}  &  $18.95 \: (0.13)$\,\% & $35.4 \: (0.3)$\,\% & -\\
\textit{zero} & $2.95 \: (0.02)$\,mm & $1.9 \: (0.01)\, ^{\circ}$ & - \\
\bottomrule
\end{tabular}
\end{table}

\begin{figure}
\centering
\includegraphics[width=\linewidth]{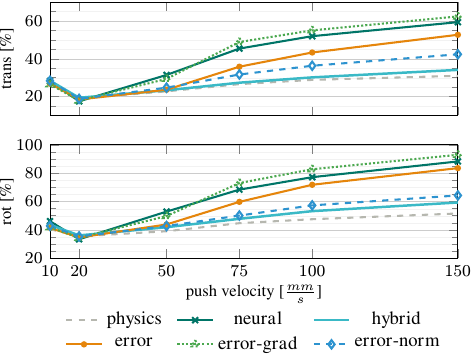}
\caption{Evaluation of the different architectures for predicting an error-correction term 
on unseen push velocities. All models were trained on push velocity 20\,$\frac{mm}
{s}$.  None of the error-prediction models is as robust as \textit{hybrid} to higher input 
velocities. \textit{Error-norm} performs best because its predicted error terms are 
independent from the push velocity. \textit{Error-grad} presumably relies more on the
error-prediction term than the other architectures and therefore performs worst
outside of the training domain.
\label{fig-error-vel}}
\end{figure}

The downside of propagating the gradients becomes apparent if we look at generalization 
to new pushing velocities: While the predictions of \textit{error} become worse with 
increasing velocity, they still remain more accurate than the predictions of \textit{neural},
as illustrated in Figure~\ref{fig-error-vel}. 
\textit{Error-grad} on the other hand performs even worse than the pure neural network. A
reason for this difference could be that \textit{error-grad} relies more strongly on the 
error-correction term than \textit{error}. This allows it to fit the training data more 
closely but at the same time impedes generalization to novel actions.

As explained before, the reason for the decline in performance when extrapolating is that 
the neural networks cannot scale their predictions correctly according to the input velocity. 
One possibility to make the error-prediction more robust to higher input velocities is the 
architecture we call \textit{error-norm}. In this model, we scale the push action to unit 
length before using it as input to the error-prediction. This makes the error-prediction 
independent of the magnitude of the action, while still giving it information about the push 
direction.
The resulting model performs only slightly worse than \textit{error} inside the training 
domain, but much better for extrapolation. It is still worse than \textit{hybrid} though,
as it cannot properly adapt the error-term to match higher velocities.

\subsubsection*{Compensation of Model Errors}
Using the error-correction term of course becomes much more interesting if the analytical
model is bad. To test how well the \textit{hybrid} and \textit{error} architectures can 
compensate for wrong models, we manipulate the friction parameter $l$ by setting it to 1.5 
or 3 times its real value. The results are shown in Table~\ref{tab-mod-fric}.

Wrong values of $l$ are especially harmful for predicting the rotation of the 
object, and both \textit{hybrid} and \textit{error} perform better than the 
\textit{physics} baseline under this condition. This shows that the \textit{hybrid} 
architecture has the ability to compensate for some errors of the analytical model by manipulating
the predicted state representation. However, while \textit{hybrid} performs
similar to \textit{error} if $l$ is only 1.5 times bigger than the correct value, 
it cannot compensate as well for larger deviations in $l$. In this case, the 
ability of \textit{error} to directly alter the output of the analytical model instead of only
manipulating its input values proves to be necessary for achieving good performance.

\begin{figure*}
\centering
\begin{subfigure}{0.245\textwidth}
\centering
\includegraphics[width=\textwidth]{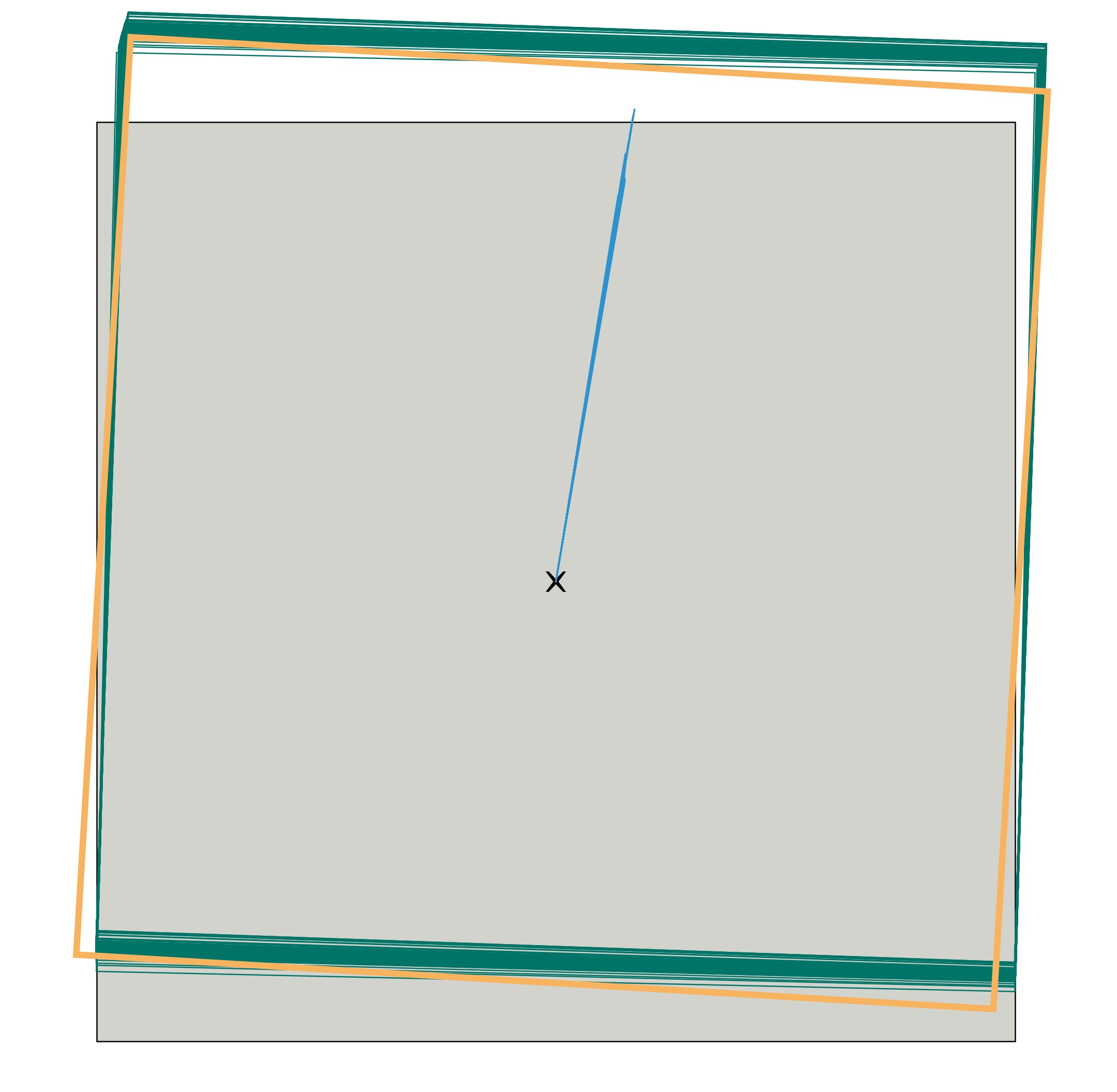}
\caption{\textit{Physics} }
\end{subfigure}
\begin{subfigure}{0.245\textwidth}
\centering
\includegraphics[width=\textwidth]{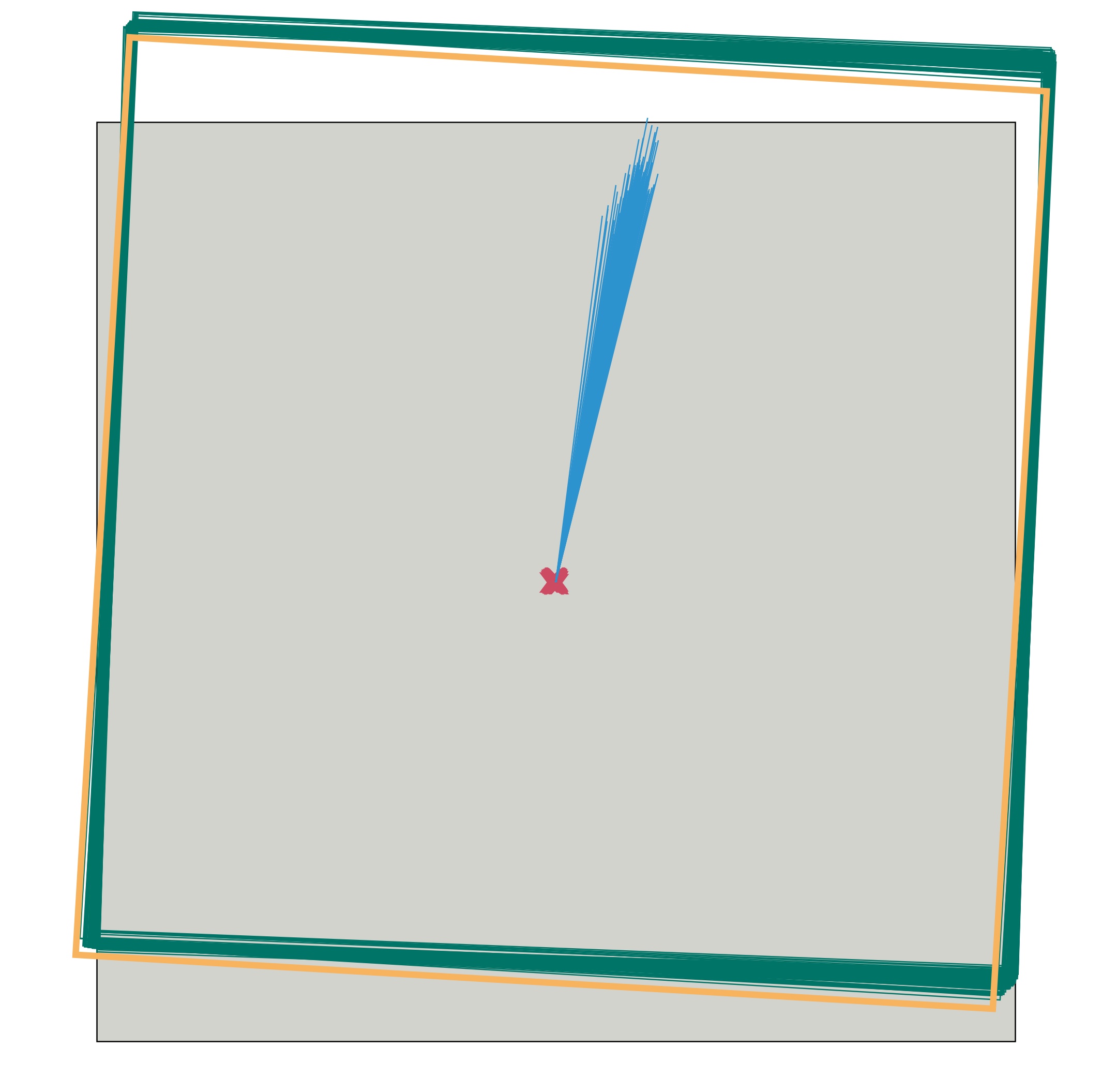}
\caption{\textit{Hybrid}}
\end{subfigure}
\begin{subfigure}{0.245\textwidth}
\includegraphics[width=\textwidth]{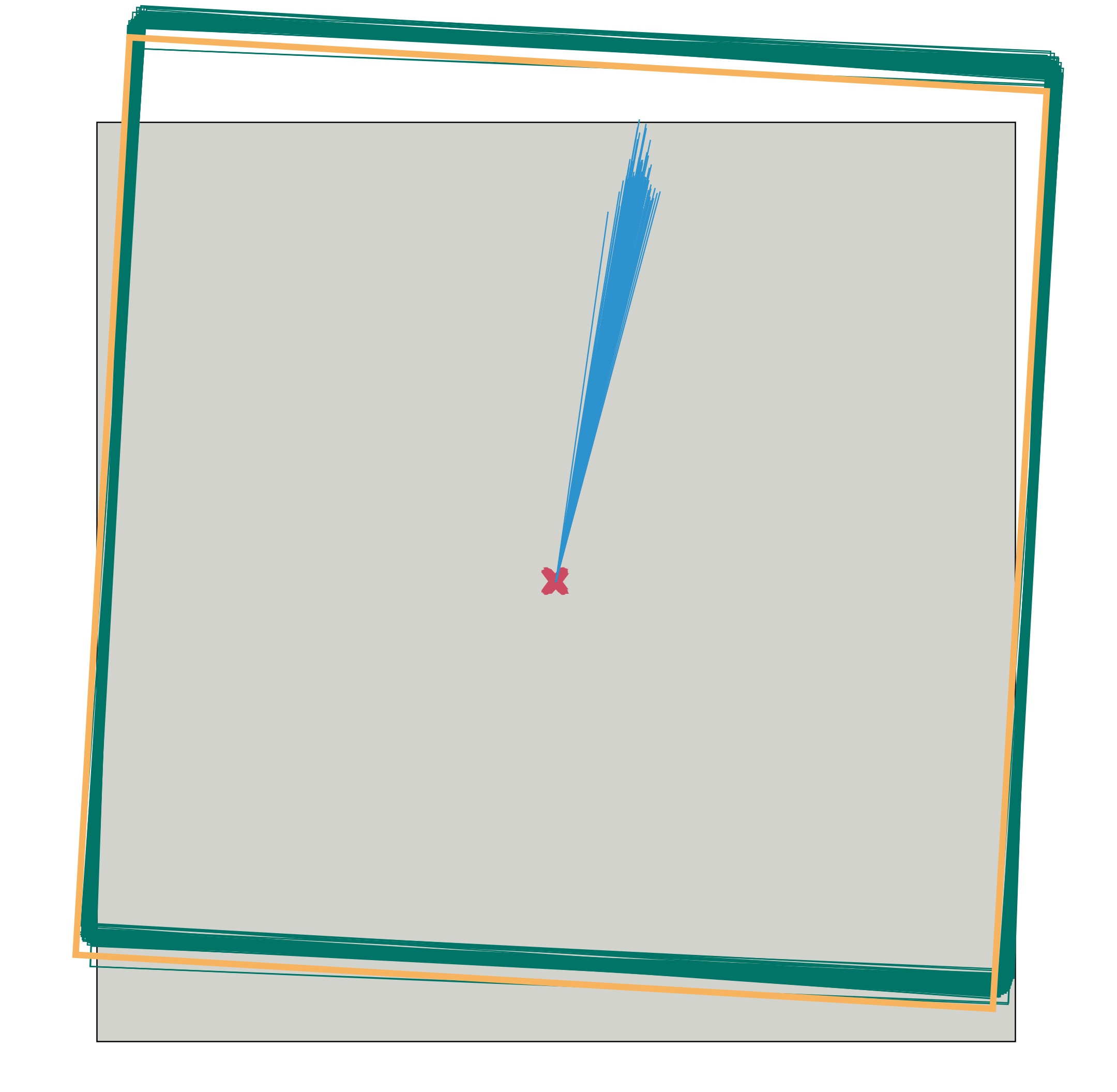}
\caption{\textit{Error}}
\end{subfigure}\\
\begin{subfigure}{0.245\textwidth}
\includegraphics[width=\textwidth]{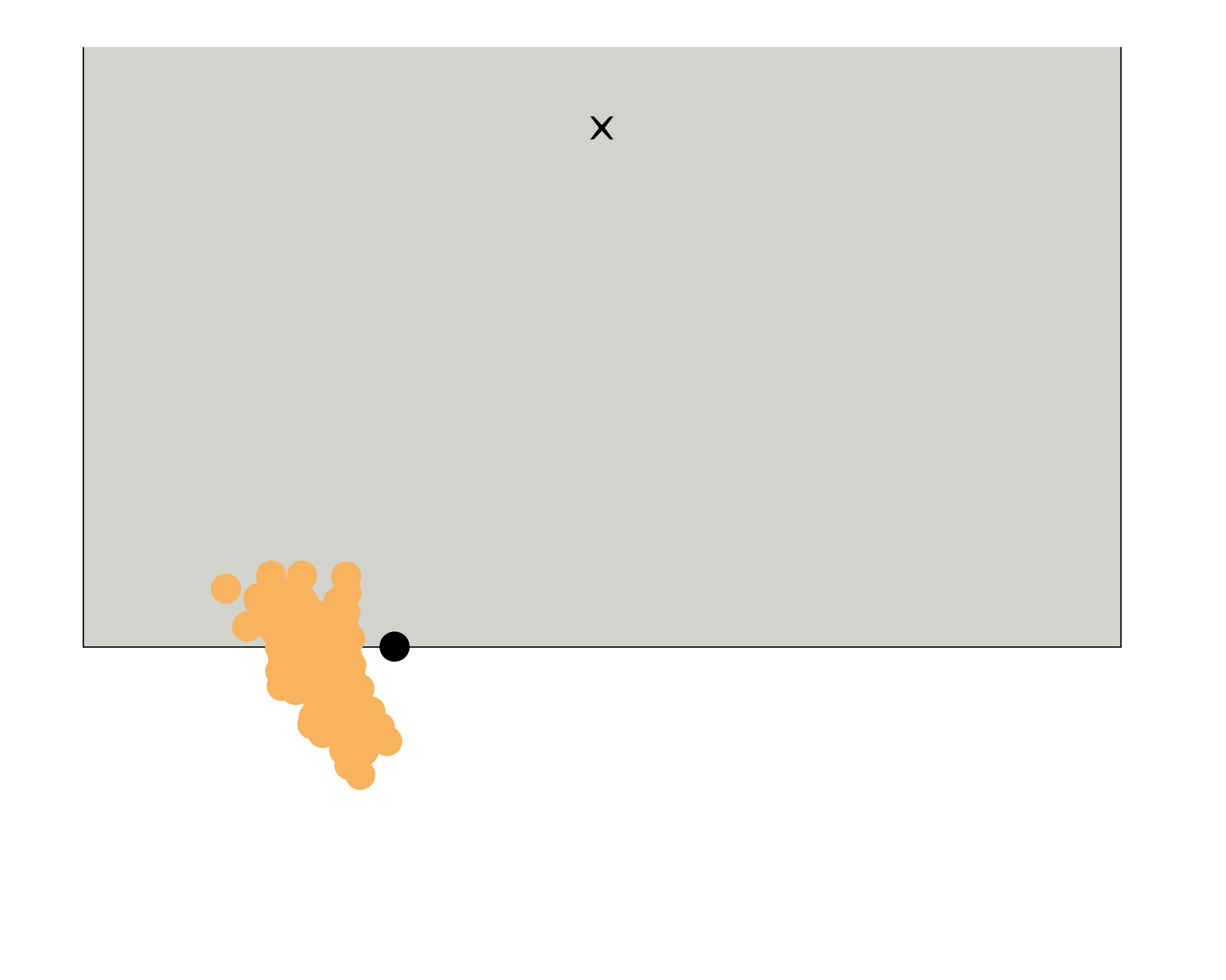}
\caption{Contact points predicted by \textit{hybrid}}
\end{subfigure}
\begin{subfigure}{0.245\textwidth}
\includegraphics[width=\textwidth]{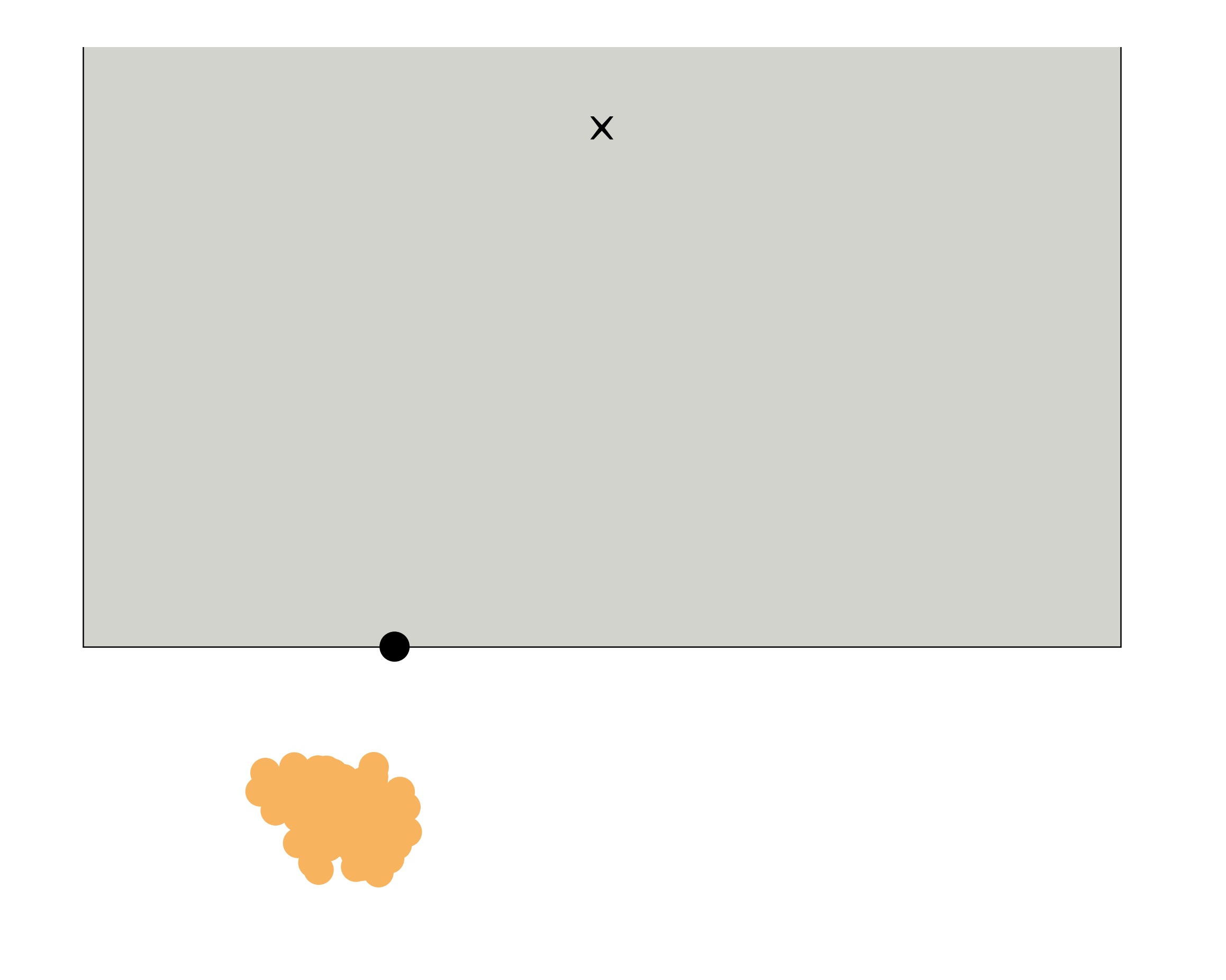}
\caption{Contact points predicted by \textit{error}}
\end{subfigure}
\begin{subfigure}{0.245\textwidth}
\includegraphics[width=\textwidth]{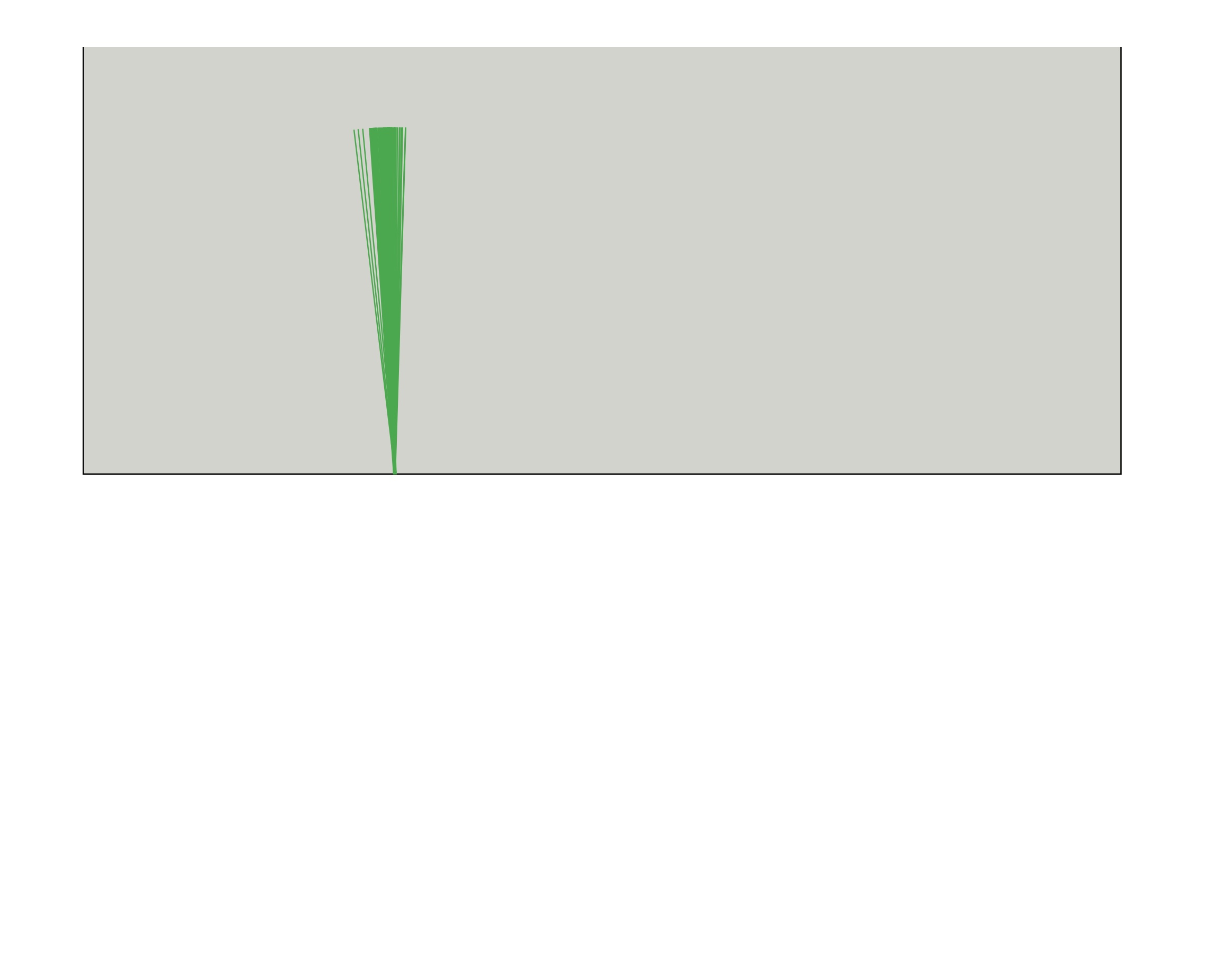}
\caption{Contact normal predicted by \textit{hybrid}}
\end{subfigure}
\begin{subfigure}{0.245\textwidth}
\includegraphics[width=\textwidth]{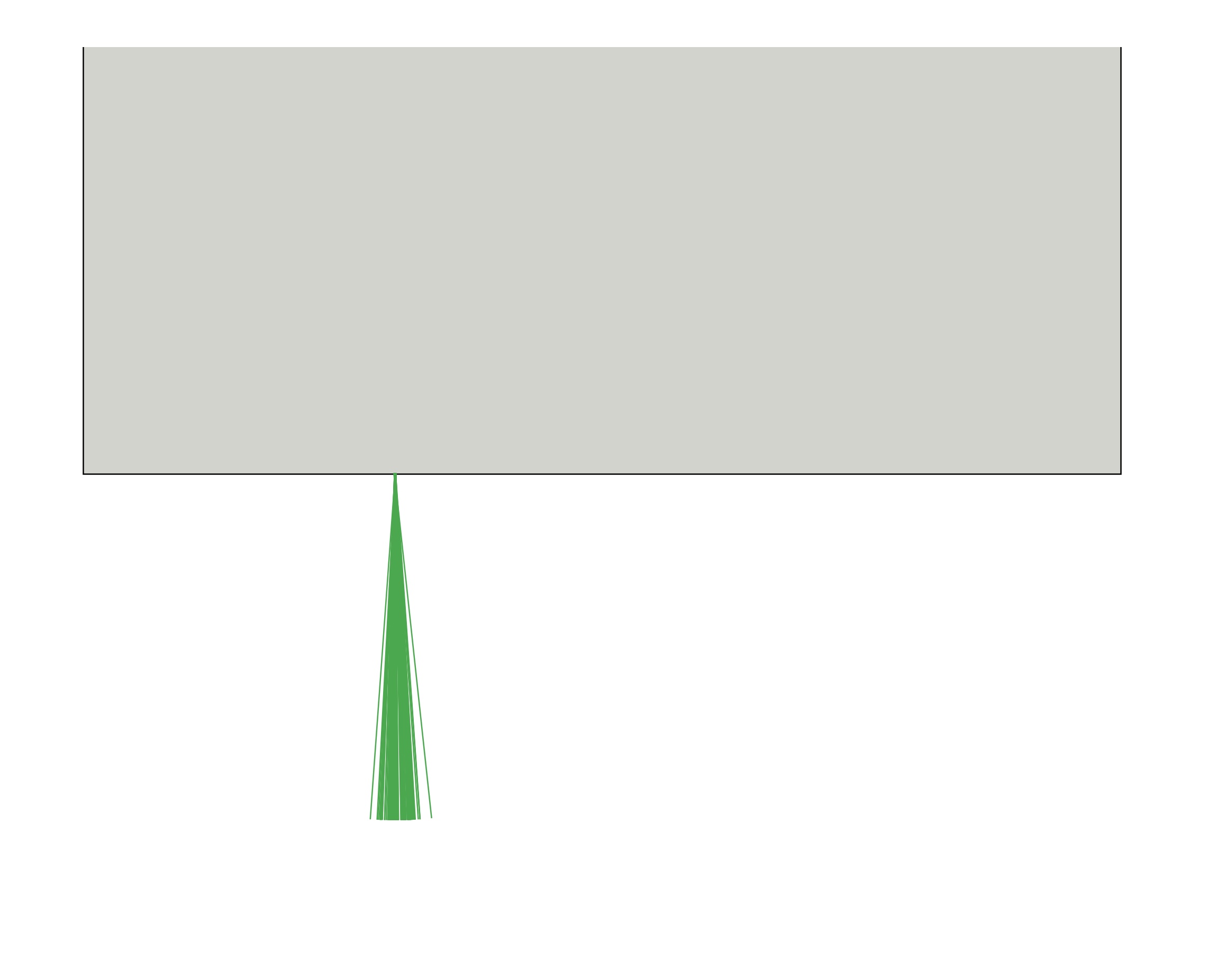}
\caption{Contact normal predicted by \textit{error}}
\end{subfigure}
\caption{Predicted movement, contact points and normals from 200 repeated pushes 
when using a wrong friction parameter ($1.5 \cdot l$). The black point marks the 
(average) ground truth contact point. Both networks compensate for the wrong 
friction parameter by predicting the contact point in a slightly wrong position, but 
the deviation from the ground truth is stronger for \textit{error}, which also flips 
the direction of the predicted normal (this is however not relevant in our implementation
of the analytical model).}
\label{fig-mod-fric}
\end{figure*}

The visualization in 
Figure~\ref{fig-mod-fric} shows that both models predicted incorrect contact points to 
counter the effect of the higher friction value.  This makes sense, since the location of the
contact point influences the tradeoff between how much the object rotates and how much it 
translates. The predictions from \textit{error} deviate farther from the ground truth values, 
which shows that the additional error-term does not prevent the model from manipulating the 
input values to the analytical model. Instead, it achieves its good results by combining both 
forms of correction. 

\begin{table}
\scriptsize
\caption{Prediction errors of \textit{physics}, \textit{hybrid} and \textit{error} when
using a manipulated friction parameter $l$. In contrast to \textit{physics}, both neural 
networks can compensate for the resulting error of the analytical model. \textit{Hybrid}
can however only modify the input values to the analytical model, while \textit{error} can 
correct the model's output directly and thus compensates the error of the analytical model much better. \label{tab-mod-fric}} 
\begin{tabular}{r  l l l} 
\toprule
 & trans  & rot   & pos [mm] \\
\midrule
\textit{$1.5 \cdot l$} &  &  &  \\
\textit{hybrid}   & $20.7 \: (0.13)$\,\% & $40.5 \: (0.32)$\,\% & $0.3 \: (0.002)$  \\
\textit{error}   & $\mathbf{19.2 \: (0.14)}$\,\% & $\mathbf{35.9 \: (0.3)}$\,\% & $\mathbf{0.25 \: (0.002)}$  \\
\textit{physics}  &  $23.9 \: (0.15)$\,\% & $46.1 \: (0.37)$\,\% & -\\
\midrule
\textit{$3 \cdot l$} &  &  &  \\
\textit{hybrid}   &$25.1 \: (0.15)$\,\% & $66.9 \: (0.45)$\,\% & $\mathbf{0.31 \: (0.002)}$  \\
\textit{error}   & $\mathbf{19.6 \: (0.13)}$\,\% & $\mathbf{37.2 \: (0.3)}$\,\% & $0.32 \: (0.002)$  \\
\textit{physics}  &  $35.6 \: (0.23)$\,\% & $80.1 \: (0.53)$\,\% & -\\
\bottomrule
\end{tabular}
\end{table}

\subsubsection*{Summary}

Adding an learned error-correction term to the hybrid approach 
improves its ability to compensate for errors in the analytical model.
It however does not prevent prediction of ``wrong'' state representations 
in such cases.
For generalization, we found it helpful to limit the error term's dependency on the
magnitude of the pushing action and to stop gradient flow from the
error to the perception module.

\section{Extension to Non-Trivial Viewpoints} \label{sec:3d}

In the previous section, we used depth images that showed a top-down view of the scene.
This simplified the perception part and allowed us to focus our experiments 
on comparing the different architectures for learning the dynamics model.
In this Section, we briefly describe what changes when we move away from the top-down 
perspective and show that the proposed hybrid approach still works well on scenes that were 
recorded from an arbitrary viewpoint.

\subsection{Challenges}\label{sec:3d-challenges}

For describing the scene geometrically, we need three coordinate frames, the
world frame, the camera frame and a frame that is attached to the object.
In our rendered scenes, the origin of the world frame is located at the centre of
the table and its $x-y$ plane is aligned with the table surface. The object
frame is attached to the centre of mass of the object. In planar pushing, we expect 
that the object only moves and rotates in the $x-y$ plane of its supporting surface
and the $x-y$ plane of object and world frame are thus aligned. The movement of
the object is described relative to the world frame.

The camera-frame is located at the sensor position and its $z$-axis 
points towards the origin of the world frame.  Depth images contain the distance
of each visible point to the sensor along the $z$ axis, i.e.\ the $z$ coordinates
of said point in camera frame.
Finally, a 3D point $(x~y~z)^T$ in camera coordinates is mapped to a
pixel $(u~v)^T$ by applying the perspective projection with focal length $f$:
\begin{align}
\begin{pmatrix}
u \\ v
\end{pmatrix} = \frac{f}{z} \begin{pmatrix}
x \\ y
\end{pmatrix} 
\end{align} 

In the special case of a top-down view, the $x-y$-plane of world and camera 
frame are aligned, such that we can apply the analytical model presented in 
Section~\ref{sec:model} in both frames. In addition, the measured depth 
($z$ in camera frame) of the table and the object are independent of their
$x$ and $y$ position. This reduces the perspective projection to a scaling 
operation with a constant factor and allowed us to predict the object movement 
directly in pixel space. It also made segmenting the object very easy, since it
is associated with one specific depth value.

Without the assumption of a top-down view, predicting object  movement 
in pixel-space becomes more challenging, since the same movement will span 
more pixels the further away the object is from the camera. In addition, 
perspective distortion now affects the shape and size of the objects in pixel space, 
which is e.g.\ relevant for localizing the object.

\subsection{Network Architecture}\label{sec:3d-architecture}

We now describe the changes we made to the architectures from Section~\ref{sec:results2d}
to adapt to new camera configurations. To be able to relate pixel coordinates to
3d coordinates, we assume that we have access to the
parameters of the camera (focal length) and the transform between camera 
and world frame.

\subsubsection{Perception}

The perception part remains mostly as it was before - we use the same architecture 
as shown in Figure~\ref{fig-perc}. 
Inspired by \citet{se3}, 
we however extend the input data from depth images to full 3d point 
clouds. These point clouds are still image-shaped and can thus be treated like normal 
images whose channels encode coordinate values instead of colour or intensity.

As our architecture estimates the position of the object $\mathbf{o}$ in pixel 
space (using spatial softmax), we use the given camera parameters and transform to
calculate the corresponding position in 3d world-coordinates. 
For this we need the $z$ coordinate (depth) of the predicted pixel location, 
which we get by interpolating between the values of the four pixels
closest to the predicted coordinates.
Due to inaccuracies in the depth-values and the projection matrix, this operation
introduces an error of about 0.3\,mm, which we however consider negligible.

\subsubsection{Prediction and Training}

\begin{figure}
\centering
\includegraphics[width=\columnwidth]{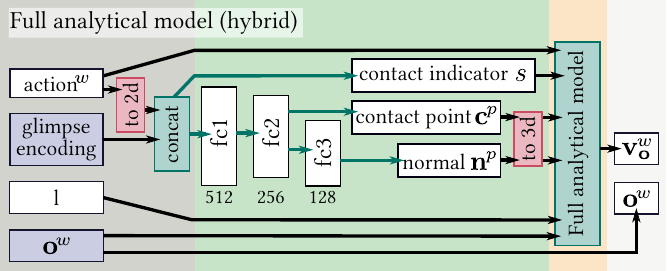}
\caption{Prediction part of variant \textit{hybrid} for non-trivial view-points.
While the analytical model operates in 3d world coordinates (indicated by $^w$),
the contact point and normal are predicted in 2d pixel space (indicated by $^p$).
Transforms from world coordinates to pixels and vice versa (red boxes) use the 
given depth values, camera parameters and transform between camera and world frame.
Please refer to Figure~\ref{fig-pred} 
for a detailed explanation of graphical elements.
%\\ 
%White and purple boxes represent tensors, where the purple color indicates
%tensors that are computed by the perception part shown in Figure~\ref{fig-perc}.
%During training, the gradient information is backpropagated through these 
%tensors to the perception part.
%\\
%Green arrows and boxes indicate network layers, whereas black arrows represent dataflow 
%without processing. In this network, all green arrows represent fully-connected layers
%and the numbers beneath their output tensors (fc) denote the number of output channels.
}\label{fig-pred-3d}
\end{figure}

If we do not use a top-down view of the scene, the question becomes relevant in 
which coordinate-frame the network should predict the contact point and the normal: pixel 
space, camera coordinates or world coordinates. We decided to continue using pixel 
coordinates since predictions in this space can be most directly related to the input 
image and the predicted feature maps. To this end, we also transform the action into
pixel space before using it (together with the glimpse encoding) as input for predicting
the contact point and normal. Since the analytical model can only be
used in the world frame where the movement of the object is limited to the $x-y$ 
plane, we finally transform the predicted contact point and normal from pixel-space 
to world-coordinates. The resulting prediction part is illustrated in Figure~\ref{fig-pred-3d}. Note that 
the overall architecture (number and size of layers) is the same as for the top-down case 
(Figure~\ref{fig-pred}).
%

%We however have to transfer the predicted contact point and normal 
%from pixel-space to world-coordinates afterwards, since the analytical model can only be
%used in the world frame where the movement of the object is limited to the $x-y$ 
%plane.

We found training the model on data from arbitrary viewpoints more 
challenging than in the top-down scenario. To facilitate the process, we modified 
the network and the training loss to treat contact prediction and velocity 
prediction separately: Instead of using the predicted contact indicator $s$ in the analytical 
model, the network now predicts $s$ as a separate output.
We interpret $s$ as the probability that the pusher is in contact with the object and 
place a cross-entropy loss on it. For the predicted velocity, we only penalize errors if 
there was (ground truth) contact. This prevents non-informative gradients to the 
velocity-prediction if the object did not move and also solves the problem that the 
network tries to compensate for wrong velocity predictions by setting the contact estimate to 
zero. 

At test-time, the predicted contact indicator is multiplied with the predicted object 
movement from the analytical model to get the final velocity prediction.

The new loss for a single training example looks as follows:
Let $\hat{s}$, $\mathbf{\hat{v}_o}$ and  $\mathbf{\hat{o}}$ denote the predicted and  
$s$, $\mathbf{v_o}$, $\mathbf{o}$ the real contact indicator, object movement and position. 
$\mathbf{w}$ are the network weights 
and $\mathbf{\nu_{o}} = [v_{ox}, v_{oy}]$  denotes linear object velocity.
\begin{gather*}
L(s, \mathbf{\hat{v}_o}, \mathbf{\hat{o}}, s, \mathbf{v_o}
, \mathbf{o}) = trans + rot + pos +  \hdots \\
 \hdots \lambda_1 \hspace{1pt} contact + \lambda_2 \sum\nolimits_{\mathbf{w}} \parallel \mathbf{w} \parallel  \\
\begin{aligned}
    trans &=  s \left\| \mathbf{\hat{\nu}_{o}} - \mathbf{\nu_{o}} \right\| \\
rot &= s \tfrac{180}{\pi}  | \omega - \hat{\omega} | \\
pos &= \parallel \mathbf{o} - \mathbf{\hat{o}} \parallel \\
contact &= -(s\log(\hat{s})+(1-s)\log(1-\hat{s}))
\end{aligned}\\
\end{gather*}

To ensure that all components of the loss are of the same magnitude, we compute
translation and position error in millimetres and rotation in degree. We set 
$\lambda_1 = 10$  and $\lambda_2 = 0.001$.

\subsection{Evaluation}\label{sec:results3d}

\begin{figure*}
\centering
\begin{subfigure}{0.3\textwidth}
\includegraphics[width=\textwidth]{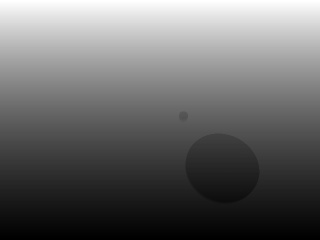}
\end{subfigure}~
\begin{subfigure}{0.3\textwidth}
\includegraphics[width=\textwidth]{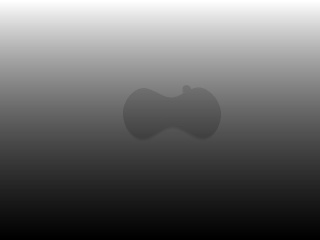}
\end{subfigure}~
\begin{subfigure}{0.3\textwidth}
\includegraphics[width=\textwidth]{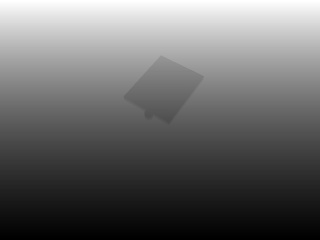}
\end{subfigure}
\caption{Example depth images recorded from a non-top-down viewpoint. The depth values 
increase towards the back of the scene and the perspective transform affects the shape of 
the objects.}\label{fig-example-image}
\end{figure*}

To show that our approach is not dependent on the top-down perspective, we train and evaluate 
it on images collected from a different viewpoint: The camera is located  at 
$\begin{pmatrix}0~-0.25~0.4\end{pmatrix}$ in world-coordinates, which corresponds to a 
more natural perspective where the viewer stands in front of the table instead of hovering 
above the table centre.
Figure~\ref{fig-example-image} shows the depth channel of three examples.

\subsubsection*{Data}

As in Section~\ref{sec:results2d}, we use a dataset that contains all objects 
from the MIT Push dataset and  all pushes with velocity $20\frac{mm}{s}$ for evaluating 
on this new viewpoint. We split it randomly into about 190k training examples and about 
37k examples for testing.

\subsubsection*{Results}

After 75k training steps, the \textit{hybrid} model trained on the new viewpoint performs 
very similar to the one from the top-down case: $19.4\,\%$ translation error and $37.2\,\%$  
error in rotation, as compared to $19.3\,\%$ and $36.1\,\%$ in the top-down case. The only 
notable difference is that the error in predicted position climbs from 0.32\,mm to 1.12\,mm. 
This can be partially explained by the inaccuracy introduced when converting from pixel 
coordinates to world coordinates. Another problem could be perspective 
distortion, as it influences 
shape and size of the objects depending on how close they are to the camera. In relation to 
the size of the object, an error of around 1\,mm is however still very small.

\subsubsection*{Summary}

The accuracy of our proposed method is not harmed by using a non-trivial camera viewpoint. 
We however require that the transform between world and camera coordinates as well
as the camera parameters are known.

\section{Conclusion and Future Work}
\label{sec:discussion}

In this paper, we considered the problem of predicting the effect of
physical interaction from raw sensory data. We compared a pure neural
network approach to a hybrid approach that uses a neural network for perception 
and an analytical model for prediction. Our test bed involved pushing of
planar objects on a surface - a non-linear,
discontinuous manipulation task for which we have both, millions of
data points and an analytical model. 

We observed two main advantages of the hybrid architecture. Compared
to the pure neural network, it significantly (i) reduces required training
data and (ii) improves generalization to novel physical
interaction. The analytical model aides generalization by limiting 
the ability of the hybrid architecture to overfit to the training data and 
by providing  multiplication operations for scaling the output according to 
the input action and contact indicator. This kind of mathematical
operation is hard to learn for fully-connected architectures and requires
many parameters and diverse training examples for covering a large value range.
The drawback of the \textit{hybrid} approach is that it cannot as easily improve
on the performance of the underlying analytical model. 

The pure neural network on the other hand can beat both, the hybrid approach and  
the analytical model (with ground truth input values) if trained on
enough data. This however only holds when we evaluate on actions
encountered during training and does not transfer to new push configurations, 
velocities or object shapes. The challenge in these cases is that the 
distribution of the training and test data differ significantly.

To enable the hybrid approach to improve more on the prediction accuracy of its
analytical model, we experimented with learning an error-correction term that
is added to the prediction of the analytical model. These \textit{error} models 
are almost as data-efficient as \textit{hybrid}
and can to some extend retain the ability to generalize to different test data 
provided by the analytical model. They however require more diversity in the training 
data than \textit{hybrid} to avoid overfitting. Our experiments with a wrong analytical model 
also showed that the \textit{error} models can compensate for errors of the model much better 
than \textit{hybrid}, which can only influence the prediction by manipulating the input values 
of the analytical model.

The last architecture, \textit{simple}, showed that combining learning and analytical 
models is not automatically guaranteed to lead to good performance. By replacing the
first stage of the analytical model with a neural network, we instead combined the 
disadvantages of both approaches: The architecture needs lots of training data and 
does not generalize well to new pushes, because it misses the part of the analytical 
model that explains the influence of the pushing action on the resulting object velocity. 
In contrast to the pure neural network, it however also cannot improve much on the performance
of the analytical model.

A limitation of the presented hybrid approach is that it may be hard to find an accurate
analytical model for some physical processes and that not all existing  
models are suitable for our approach, as they e.g.\ need to be differentiable everywhere. 
Especially the switching dynamics encountered when the contact situation
changes proved to be challenging and more work needs to be done in this 
direction. If no analytical model is available, learning the predictive model with a 
neural network is still a very good option.

In perception on the other hand, the strengths of neural networks can 
be well exploited to extract the input parameters of the analytical 
model from raw sensory data. By training end-to-end through a given model, 
we can avoid the effort of labeling data with the ground truth state. 
Our experiments also showed that training end-to-end allows the hybrid models 
to compensate for smaller errors in the analytical model by adjusting the 
predicted input values. 

Using the state representation of the analytical model for the \textit{hybrid} 
architecture has the advantage that the predictions of the network can be visualized 
and interpreted. This is not easily possible for the intermediate representations
learned in  the pure neural network. Our results however suggest that the 
pure neural network benefits from being free to chose its own state representation, as 
learning the dynamics model from the ground truth state representation (\textit{neural dyn}) 
lead to worse prediction results.

In the future, we want to extend our work to more complex
perception problems, like training on RGB images or scenes with multiple objects. An
interesting question is if perception on  point clouds could be facilitated by using 
methods like pointnet++~\citep{pointnet++} that are specifically designed for this type of 
input data instead of treating the point cloud like a normal image. 

A logical next step is also using our hybrid models for predicting more than one step into 
the future. Working on sequences makes it for example possible
to guide learning by enforcing constraints like temporal consistency or to exploit temporal
cues like optical flow. The model could also be used in a filtering scenario to track the 
state of an object and at the same time infer latent variables of the system like the 
friction coefficients. Similarly, we can use the learned model to plan and execute
robot actions using model predictive control.

\section*{Acknowledgments}
This work was supported by the Max Planck Society. The authors thank the International 
Max Planck Research School for Intelligent Systems (IMPRS-IS) for supporting Alina Kloss.

\theendnotes
\bibliographystyle{abbrvnat}
\bibliography{paper}

\begin{thebibliography}{29}
\providecommand{\natexlab}[1]{#1}
\providecommand{\url}[1]{\texttt{#1}}
\expandafter\ifx\csname urlstyle\endcsname\relax
  \providecommand{\doi}[1]{doi: #1}\else
  \providecommand{\doi}{doi: \begingroup \urlstyle{rm}\Url}\fi

\bibitem[Abadi et~al.(2015)]{tensorflow}
M.~Abadi et~al.
\newblock {TensorFlow}: Large-scale machine learning on heterogeneous systems,
  2015.
\newblock Software available from tensorflow.org.

\bibitem[Agrawal et~al.(2016)Agrawal, Nair, Abbeel, Malik, and Levine]{poke}
P.~Agrawal, A.~V. Nair, P.~Abbeel, J.~Malik, and S.~Levine.
\newblock Learning to poke by poking: Experiential learning of intuitive
  physics.
\newblock In \emph{Advances in Neural Inform. Process. Syst.}, pages
  5074--5082, 2016.

\bibitem[Bauza and Rodriguez(2017)]{bauza}
M.~Bauza and A.~Rodriguez.
\newblock A probabilistic data-driven model for planar pushing.
\newblock In \emph{2017 IEEE International Conference on Robotics and
  Automation (ICRA)}, pages 3008--3015, May 2017.
\newblock \doi{10.1109/ICRA.2017.7989345}.

\bibitem[Belter et~al.(2014)Belter, Kopicki, Zurek, and Wyatt]{belter2014}
D.~Belter, M.~Kopicki, S.~Zurek, and J.~Wyatt.
\newblock Kinematically optimised predictions of object motion.
\newblock In \emph{Intelligent Robots and Systems (IROS 2014), 2014 IEEE/RSJ
  International Conference on}, pages 4422--4427. IEEE, 2014.

\bibitem[Byravan and Fox(2017)]{se3}
A.~Byravan and D.~Fox.
\newblock Se3-nets: Learning rigid body motion using deep neural networks.
\newblock In \emph{Robotics and Automation (ICRA), 2017 IEEE Int. Conf. on},
  pages 173--180. IEEE, 2017.

\bibitem[Byravan et~al.(2017)Byravan, Leeb, Meier, and Fox]{se3pose}
A.~Byravan, F.~Leeb, F.~Meier, and D.~Fox.
\newblock Se3-pose-nets: Structured deep dynamics models for visuomotor
  planning and control.
\newblock \emph{to appear at Robotics and Automation (ICRA), 2018 IEEE Int.
  Conf. on}, abs/1710.00489, 2017.

\bibitem[Degrave et~al.(2016)Degrave, Hermans, Dambre, and
  Wyffels]{differentiabe_pysics_engine}
J.~Degrave, M.~Hermans, J.~Dambre, and F.~Wyffels.
\newblock A differentiable physics engine for deep learning in robotics.
\newblock \emph{CoRR}, abs/1611.01652, 2016.

\bibitem[Finn et~al.(2016)Finn, Goodfellow, and Levine]{NIPS2016_6161}
C.~Finn, I.~Goodfellow, and S.~Levine.
\newblock Unsupervised learning for physical interaction through video
  prediction.
\newblock In \emph{Advances in Neural Inform. Process. Syst. 29}, pages 64--72.
  2016.

\bibitem[Goyal et~al.(1991)Goyal, Ruina, and Papadopoulos]{goyal}
S.~Goyal, A.~Ruina, and J.~Papadopoulos.
\newblock Planar sliding with dry friction part 1. limit surface and moment
  function.
\newblock \emph{Wear}, 143\penalty0 (2):\penalty0 307 -- 330, 1991.
\newblock ISSN 0043-1648.
\newblock \doi{https://doi.org/10.1016/0043-1648(91)90104-3}.

\bibitem[Hong~Lee and Cutkosky(1991)]{lee-cutkosky}
S.~Hong~Lee and M.~Cutkosky.
\newblock Fixture planning with friction.
\newblock \emph{Journal of Engineering for Industry}, 113, 08 1991.
\newblock URL
  \url{http://manufacturingscience.asmedigitalcollection.asme.org/article.aspx?articleid=1447458}.

\bibitem[Howe and Cutkosky(1996)]{howe-cutkosky}
R.~D. Howe and M.~R. Cutkosky.
\newblock Practical force-motion models for sliding manipulation.
\newblock \emph{The International Journal of Robotics Research}, 15\penalty0
  (6):\penalty0 557--572, 1996.
\newblock \doi{10.1177/027836499601500603}.

\bibitem[Ioffe and Szegedy(2015)]{batch_norm}
S.~Ioffe and C.~Szegedy.
\newblock Batch normalization: Accelerating deep network training by reducing
  internal covariate shift.
\newblock In \emph{Proc. 32nd Int. Conf. on Machine Learning}, volume~37, pages
  448--456, 07--09 Jul 2015.

\bibitem[Jonschkowski and Brock(2015)]{robotic_priors}
R.~Jonschkowski and O.~Brock.
\newblock Learning state representations with robotic priors.
\newblock \emph{Autonomous Robots}, 39\penalty0 (3):\penalty0 407--428, Oct
  2015.
\newblock ISSN 1573-7527.
\newblock \doi{10.1007/s10514-015-9459-7}.

\bibitem[Kingma and Ba(2014)]{adam}
D.~Kingma and J.~Ba.
\newblock Adam: A method for stochastic optimization.
\newblock \emph{arXiv preprint arXiv:1412.6980}, 2014.

\bibitem[Kopicki et~al.(2017)Kopicki, Zurek, Stolkin, Moerwald, and
  Wyatt]{kopicki2017}
M.~Kopicki, S.~Zurek, R.~Stolkin, T.~Moerwald, and J.~L. Wyatt.
\newblock Learning modular and transferable forward models of the motions of
  push manipulated objects.
\newblock \emph{Autonomous Robots}, 41\penalty0 (5):\penalty0 1061--1082, 6
  2017.
\newblock \doi{10.1007/s10514-016-9571-3}.

\bibitem[Levine et~al.(2016)Levine, Finn, Darrell, and Abbeel]{rf2}
S.~Levine, C.~Finn, T.~Darrell, and P.~Abbeel.
\newblock End-to-end training of deep visuomotor policies.
\newblock \emph{J. Mach. Learning Research}, 17\penalty0 (39):\penalty0 1--40,
  2016.

\bibitem[Lillicrap et~al.(2015)Lillicrap, Hunt, Pritzel, Heess, Erez, Tassa,
  Silver, and Wierstra]{rf1}
T.~P. Lillicrap, J.~J. Hunt, A.~Pritzel, N.~Heess, T.~Erez, Y.~Tassa,
  D.~Silver, and D.~Wierstra.
\newblock Continuous control with deep reinforcement learning.
\newblock \emph{CoRR}, abs/1509.02971, 2015.

\bibitem[Lynch et~al.(1992)Lynch, Maekawa, and Tanie]{model}
K.~M. Lynch, H.~Maekawa, and K.~Tanie.
\newblock Manipulation and active sensing by pushing using tactile feedback.
\newblock In \emph{Proc. IEEE/RSJ Int. Conf. Intelligent Robots and Systems},
  volume~1, pages 416--421, Jul 1992.
\newblock \doi{10.1109/IROS.1992.587370}.

\bibitem[Martius and Lampert(2016)]{eql}
G.~Martius and C.~H. Lampert.
\newblock Extrapolation and learning equations.
\newblock \emph{CoRR}, abs/1610.02995, 2016.
\newblock URL \url{http://arxiv.org/abs/1610.02995}.

\bibitem[Mason(1986)]{mason}
M.~T. Mason.
\newblock Mechanics and planning of manipulator pushing operations.
\newblock \emph{The International Journal of Robotics Research}, 5\penalty0
  (3):\penalty0 53--71, 1986.

\bibitem[Meri{\c{c}}li et~al.(2015)Meri{\c{c}}li, Veloso, and
  Ak{\i}n]{mericcli}
T.~Meri{\c{c}}li, M.~Veloso, and H.~L. Ak{\i}n.
\newblock Push-manipulation of complex passive mobile objects using
  experimentally acquired motion models.
\newblock \emph{Autonomous Robots}, 38\penalty0 (3):\penalty0 317--329, 2015.

\bibitem[Nguyen-Tuong and Peters(2010)]{gp}
D.~Nguyen-Tuong and J.~Peters.
\newblock Using model knowledge for learning inverse dynamics.
\newblock In \emph{Robotics and {Automation} ({ICRA}), 2010 {IEEE} {Int.}
  {Conf.} on}, pages 2677--2682. IEEE, 2010.

\bibitem[Qi et~al.(2017)Qi, Yi, Su, and Guibas]{pointnet++}
C.~R. Qi, L.~Yi, H.~Su, and L.~J. Guibas.
\newblock Pointnet++: Deep hierarchical feature learning on point sets in a
  metric space.
\newblock In \emph{Advances in Neural Inform. Process. Syst.}, pages
  5105--5114, 2017.

\bibitem[{Watters} et~al.(2017){Watters}, {Tacchetti}, {Weber}, {Pascanu},
  {Battaglia}, and {Zoran}]{visual_interaction}
N.~{Watters}, A.~{Tacchetti}, T.~{Weber}, R.~{Pascanu}, P.~{Battaglia}, and
  D.~{Zoran}.
\newblock {Visual Interaction Networks}.
\newblock \emph{ArXiv e-prints}, June 2017.

\bibitem[Wu et~al.(2017)Wu, Lu, Kohli, Freeman, and Tenenbaum]{deanimation}
J.~Wu, E.~Lu, P.~Kohli, B.~Freeman, and J.~Tenenbaum.
\newblock Learning to see physics via visual de-animation.
\newblock In I.~Guyon, U.~V. Luxburg, S.~Bengio, H.~Wallach, R.~Fergus,
  S.~Vishwanathan, and R.~Garnett, editors, \emph{Advances in Neural
  Information Processing Systems 30}, pages 152--163. Curran Associates, Inc.,
  2017.

\bibitem[Yu et~al.(2016)Yu, Bauza, Fazeli, and Rodriguez]{data}
K.~T. Yu, M.~Bauza, N.~Fazeli, and A.~Rodriguez.
\newblock More than a million ways to be pushed. a high-fidelity experimental
  dataset of planar pushing.
\newblock In \emph{2016 IEEE/RSJ Int. Conf. Intelligent Robots and Systems
  (IROS)}, pages 30--37, Oct 2016.
\newblock \doi{10.1109/IROS.2016.7758091}.

\bibitem[Zhang et~al.(2016)Zhang, Bengio, Hardt, Recht, and
  Vinyals]{generalization}
C.~Zhang, S.~Bengio, M.~Hardt, B.~Recht, and O.~Vinyals.
\newblock Understanding deep learning requires rethinking generalization.
\newblock \emph{CoRR}, abs/1611.03530, 2016.

\bibitem[Zhang and Trinkle(2012)]{6225125}
L.~Zhang and J.~C. Trinkle.
\newblock The application of particle filtering to grasping acquisition with
  visual occlusion and tactile sensing.
\newblock In \emph{2012 IEEE Int. Conf. on Robotics and Automation}, pages
  3805--3812, May 2012.

\bibitem[Zhou et~al.(2016)Zhou, Paolini, Bagnell, and Mason]{zhou}
J.~Zhou, R.~Paolini, J.~A. Bagnell, and M.~T. Mason.
\newblock A convex polynomial force-motion model for planar sliding:
  Identification and application.
\newblock In \emph{Robotics and Automation (ICRA), 2016 IEEE International
  Conference on}, pages 372--377. IEEE, 2016.

\end{thebibliography}

\end{document}